\documentclass[10pt,twocolumn,letterpaper]{article}

\usepackage{iccv}
\usepackage{times}
\usepackage{epsfig}
\usepackage{graphicx}
\usepackage{amsmath}
\usepackage{amssymb}
\usepackage{comment}
\usepackage{enumitem}

\usepackage[breaklinks=true,bookmarks=false,colorlinks]{hyperref}

\usepackage[font=small, labelsep=period]{caption}

\iccvfinalcopy %

\ificcvfinal\fi

\begin{document}

\title{Non-Rigid Neural Radiance Fields: Reconstruction and Novel View Synthesis of a Dynamic Scene From Monocular Video}

\author{\hspace{2em}Edgar Tretschk\\
\hspace{2em}MPI for Informatics, SIC
\and
\hspace{3em}Ayush Tewari\\
\hspace{2em}MPI for Informatics, SIC
\and
\hspace{2em}Vladislav Golyanik\\
\hspace{2em}MPI for Informatics, SIC
\and
Michael Zollh\"ofer\\
Facebook Reality Labs Research
\and
Christoph Lassner\\
Facebook Reality Labs Research
\and
Christian Theobalt\\
MPI for Informatics, SIC
}

\twocolumn[{ 
\renewcommand\twocolumn[1][]{#1} 
\maketitle 
\begin{center}
    \includegraphics[width=0.93\textwidth]{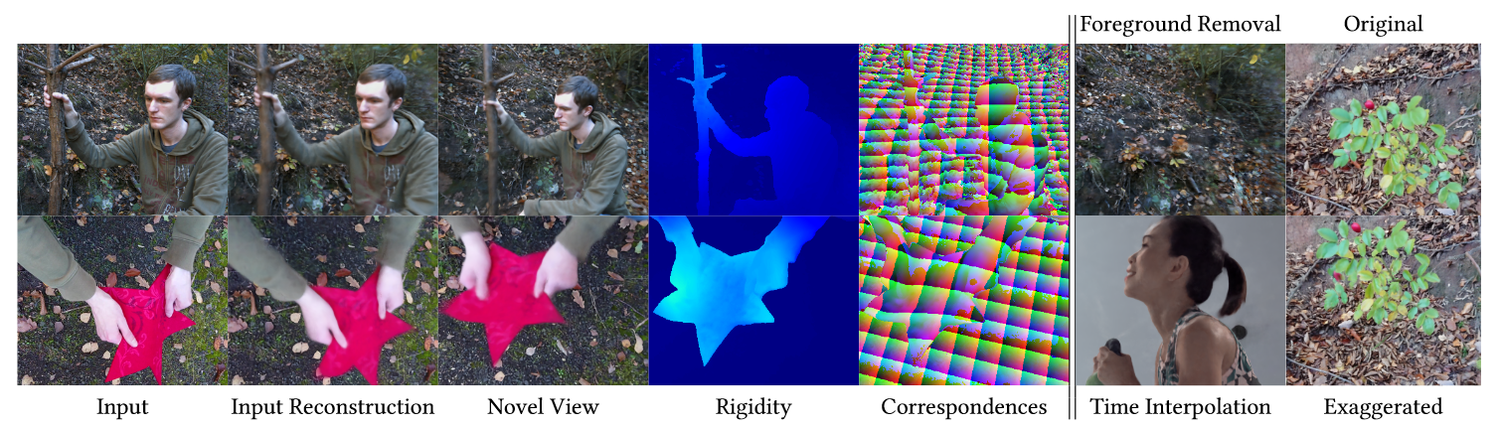}
    \captionof{figure}{
    Given a monocular image sequence, NR-NeRF reconstructs a single canonical neural radiance field to represent geometry and appearance, and a per-time-step deformation field. 
    We can render the scene into a novel spatio-temporal camera trajectory that significantly differs from the input trajectory. 
    NR-NeRF also learns rigidity scores and correspondences without direct supervision on either. 
    We can use the rigidity scores to remove the foreground, we can supersample along the time dimension, and we can exaggerate or dampen motion.
    }
    \label{fig:rigidity_correspondences} 
\end{center} 
}] 

\maketitle
\ificcvfinal\thispagestyle{empty}\fi

\begin{abstract}
   We present Non-Rigid Neural Radiance Fields (NR-NeRF), a reconstruction and novel view synthesis approach for general non-rigid dynamic scenes.
Our approach takes RGB images of a dynamic scene as input (e.g., from a monocular video recording), and creates a high-quality space-time geometry and appearance representation.
We show that a single handheld consumer-grade camera is sufficient to synthesize sophisticated renderings of a dynamic scene from novel virtual camera views, e.g. a `bullet-time' video effect. 
NR-NeRF disentangles the dynamic scene into a canonical volume and its deformation. 
Scene deformation is implemented as ray bending, where straight rays are deformed non-rigidly. %
We also propose a novel rigidity network to better constrain rigid regions of the scene, leading to more stable results.
The ray bending and rigidity network are trained without explicit supervision.
Our formulation enables dense correspondence estimation across views and time, and compelling video editing applications such as motion exaggeration.
Our code will be open sourced.

\end{abstract}

\section{Introduction} 

Free viewpoint rendering is a well-studied problem due to its wide range of applications in movies and virtual/augmented reality \cite{smolic20063d,collet2015high,miller2005interactive}.
In this work, we are interested in dynamic scenes, which change over time, from novel user-controlled viewpoints.
Traditionally, multi-view recordings are required for free viewpoint  rendering of dynamic scenes \cite{Zhang2003, Tung2009, Oswald2014}.
However, such multi-view captures are expensive and cumbersome. %
We would like to enable the setting in which a casual user records a dynamic scene with a single, moving consumer-grade camera. 
Access to only a monocular video of the deforming scene leads to a severely under-constrained problem.  
Most existing approaches thus limit themselves to a single object category, such as the human body~\cite{Habermann:2019:LRH:3313807.3311970,xiang2019monocular,VIBE:CVPR:2020} or face~\cite{egger_3DMM_survey}. 
Some approaches allow for the reconstruction of general non-rigid objects~\cite{zollhofer2018state,Garg2013,Kumar2018,Sidhu2020}, but most methods only reconstruct the geometry without the appearance of the objects in the scene. 
In contrast, our objective is to reconstruct a general dynamic scene, including its appearance, such that it can be rendered from novel spatio-temporal viewpoints.

Recent neural rendering approaches have shown impressive novel-view synthesis of general static scenes from multi-view input~\cite{tewari2020state}. 
These approaches represent scenes using trained neural networks and rely on less constraints about the type of scene, compared to traditional approaches. 
The closest prior work to our method is NeRF~\cite{mildenhall2020nerf}, which learns a continuous volume of the scene encoded in a neural network using multiple camera views. %
However, NeRF assumes the scene to be static. %
Neural Volumes~\cite{lombardi2019neural} is another closely related approach that uses multiple views of a deforming scene to enable free viewpoint rendering. 
However, it uses a fixed-size voxel grid to represent the reconstruction of the scene, restricting the resolution. 
In addition, it requires multi-view input for training, which limits the applicability to in-the-wild outdoor settings or existing monocular footage. 
Our new neural rendering approach instead targets the more challenging setting of using just a monocular video of a general dynamic scene.
Due to the non-rigidity, 
each image of the video records a different, deformed state of the scene, violating the constraints of standard neural rendering approaches.
Our approach disentangles the observations in any image into a canonical scene and its deformations, without direct supervision on either.

We tackle this problem using several innovations. 
We represent the non-rigid scene by %
two components: (1) a canonical neural radiance field for capturing geometry and appearance and (2) the scene deformation field. 
The canonical volume is a static representation of the scene encoded as a Multi-Layered Perceptron (MLP), which is not directly supervised. 
This volume is deformed into each individual image using the estimated scene deformation.
Specifically, the scene deformation is implemented as ray bending, where straight camera rays can %
deform non-rigidly.
The ray bending is modeled using an MLP that takes point samples on the ray as well as a latent code for each image as input.
Both the ray bending and the canonical scene MLPs are jointly trained using the monocular observations. 
Since the ray bending MLP deforms the entire space independent of camera parameters, we can render the deforming volume from static or time-varying novel viewpoints after training. 

\begin{figure}
  \centering
  \includegraphics[width=0.46\textwidth]{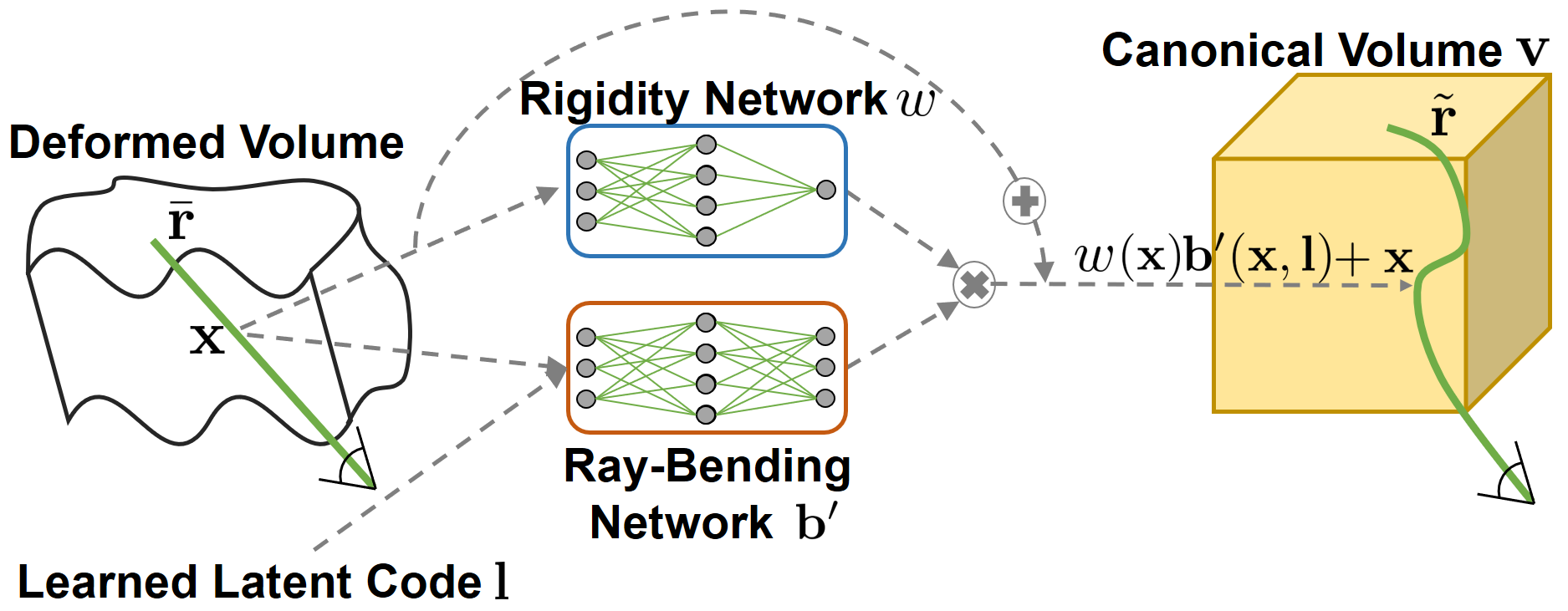}%
  \caption{
  We bend straight rays $\bar{\mathbf{r}}$ from the deformed volume using a deformation-dependent ray-bending network $\mathbf{b}'$ and a deformation-independent rigidity network $w$ into a single static canonical neural radiance field volume $\mathbf{v}$.
  }
  \label{fig:pipeline}
\end{figure}
The ray bending MLP disentangles the geometry of the scene from the scene deformations. 
The disentanglement is an underconstrained problem, which we tackle with further innovations. 
Our method assigns a rigidity score to every point in the canonical volume, which allows for the deformations to not affect the static regions in the scene. 
This rigidity component is jointly learned without any direct supervision. 
We also introduce multiple regularizers as additional soft-constraints:
A regularizer on the deformation magnitude of the \textit{visible} deformations encourages only sparse deformations of the volume, and thus helps to constrain the canonical volume. 
An additional divergence regularizer preserves the local shape, thereby constraining the representation of \textit{hidden} (partially occluded) regions that are not visible throughout the full video.

Our results show high-fidelity reconstruction and novel view synthesis for a wide range of non-rigid scenes.
Fig.~\ref{fig:pipeline} contains an overview of our method.
To summarize, our main technical \textbf{contributions} are as follows: 
\begin{itemize}[leftmargin=*]
    \itemsep0em 
    \item A free viewpoint rendering method, NR-NeRF, that only requires a monocular video of the dynamic scene %
    (Sec.~\ref{sec:method}). 
    The spatio-temporal camera trajectory for test-time novel view synthesis can differ significantly from the trajectory of the input video. 
    Moreover, we can extract dense correspondences relating arbitrary (input or novel) frames.
    \item A rigidity network which can segment the scene into non-rigid foreground and rigid background without being  directly supervised (Sec.~\ref{ssec:deformation_model}). 
    \item Regularizers on the estimated deformations which constrain the problem by encouraging small volume preserving deformations  (Sec.~\ref{subs:losses}).
    \item Several extensions for handling of view dependence and multi-view data, and applications of our technique for simple scene editing %
    (Secs.~\ref{sec:method}).
\end{itemize}
We compare NR-NeRF to several methods for neural novel  view rendering %
(Sec.~\ref{sec:results}). 
See our supplementary video for visualizations and  Sec.~\ref{sec:limitations} for a discussion.

\section{Related Work}\label{sec:related_work}

\paragraph{4D Reconstruction and Novel Viewpoint Rendering.}%
Early methods for image-based novel and free-viewpoint rendering combined traditional concepts of multi-view camera geometry, explicit vision-based 3D shape and appearance reconstruction, and classical computer graphics or image-based rendering. 
These methods are based on light fields \cite{Levoy1996, Gortler1996, Buehler2001}, multi-view stereo to capture dense depth maps \cite{Zhang2003}, layered depth images \cite{Shade1998}, or representations using 3D point clouds \cite{Agarwal2009C_ACM, Liu2010TVCG, schonberger2016structure}, meshes \cite{Matsuyama2004, Tung2009} or surfels \cite{Pfister2000, Carceroni2002, Waschbuesch2005} for dynamic scenes.
Passive geometry capture often leads to artifacts in scenes with severe occlusions and view-dependent appearance.
Also, capturing temporally coherent representations in this way is challenging.

More recently, the combination of multi-view stereo with fusion algorithms integrating implicit geometry over short time windows lead to improved results and short-term temporal coherence \cite{Dou2016, OrtsEscolano2016, Guo2017}. 
By using active depth cameras and such fusion-type reconstruction, dynamic scene capture and novel viewpoint rendering  from a low number of cameras or a single camera were shown \cite{Yu2017, DoubleFusion2018, Huang2018, Yu2019}. 
Several algorithms use variants of shape-from-silhouette to approximate real scene geometry, such as visual hull  reconstruction or visual hulls improved via multi-view photo-consistency in \cite{Kutulakos2000, Starck2006}.
While reconstruction is fast and feasible with fewer cameras, the coarse approximate geometry introduces rendering artifacts, and the reconstruction is usually limited to the separable foreground. 
Accurate and temporally coherent geometry is hard to capture in this way \cite{Cagniart2010FreeformMT, Cagniart2010}. 
Some approaches use 3D templates and combine vision-based reconstruction with appearance modelling to enable free-viewpoint video relighting, e.g., by estimating reflectance models under general lighting or under controlled light stage illumination \cite{Theobalt2007, Li2013, Nagano2015, Guo2019}. 

The progress in RGB-D sensors has enabled depth map capture from a single camera. 
Such sensors can been used for 3D reconstruction and completion of rigid environments \cite{Newcombe2011} and non-rigid objects \cite{Zollhoefer2014, Newcombe2015, Innmann2016, Slavcheva2017}. 
Other method classes allow capturing deformable geometry from sets of monocular views. 
Dense non-rigid structure from motion requires dense point tracks over input images, which are then factorized into camera poses and non-rigid 3D states per view \cite{Garg2013, Kumar2018, Sidhu2020}. 
The correspondences are usually obtained with dense optical flow methods, which makes them prone to occlusions and inaccuracies, and which can have a detrimental effect on the reconstructions. 
Monocular template-based methods do not assume dense matches and rely on a known 3D state of a deformable object (a 3D template), which is then tracked across time \cite{Perriollat2011, Ngo2015, Yu2015, Xu2018MHP}, or a training dataset with multiple object states \cite{Golyanik2018, Tretschk2020DEMEA}. 
Obtaining templates for complex objects and scenes is  often non-trivial and requires specialized setups.

In contrast, our approach avoids explicit image-based 3D reconstruction. 
Moreover, we support arbitrary backgrounds whereas the discussed methods for monocular 3D reconstruction of deformable objects ignore it. 
Our approach enables free-viewpoint rendering of general deformable scenes with multiple objects and complex deformations with high visual fidelity, 
and yet does not rely on templates, 2D correspondences and multi-view setups.

\noindent\textbf{Neural Scene Representations and Neural Rendering.} %
An emerging algorithm class uses neural networks to augment or replace established graphics and vision concepts for reconstruction and novel-view rendering. 
Most recent work is designed for static scenes \cite{Hedman2018, eslami2018neural, NguyenPhuoc2018, Meshry2019, flynn2019deepview, Sitzmann2019DV, Sitzmann2019, Mildenhall2020, Riegler2020}; methods for dynamic scenes are in their infancy. 

Several approaches address related problems to ours, such as generating images of humans in new poses \cite{Balakrishnan2018, Ma18, Neverova2018, Sarkar2020} or body reenactment from monocular videos  \cite{Chan2019dance}.
Other methods combine explicit dynamic scene reconstruction and traditional graphics rendering with neural re-rendering  \cite{MartinBrualla2018, Kim2018, Kim2019Neural, Yoon2020}. 
Shysheya~\emph{et al.}~\cite{Shysheya2019} proposed a neural rendering approach for human avatars with texture warping. 
Zhu~\emph{et al.}~\cite{Zhu2018extrapolation} leverage geometric constraints and optical flow for synthesizing novel views of humans from a single image. 
Thies~\emph{et al.}~\cite{Thies2019} combine neural textures with the classical graphics pipeline for novel view synthesis of static objects and monocular video re-rendering.
Neural Volumes \cite{lombardi2019neural} learn object models which can be animated and rendered from novel views, given multi-view video data. 
In contrast to all these methods, 
we require only a set of monocular views of a non-rigid scene as input and are able to render the scene from novel views. %

\noindent\textbf{Non-Peer-Reviewed Reports.} 
Since the intersection of neural scene representations and volumetric rendering has recently become a very active area of research with quickly evolving progress, several methods for dynamic settings have been proposed concurrently to ours. 
We mention them only for completeness since they are not peer-reviewed and thus do not constitute prior work. 
Some methods extend neural radiance fields to deforming faces \cite{wang2020learning,Gao2020,Gafni2020,raj2021pva}. 
Others focus on moving human bodies \cite{Weng2020,peng2020neural,su2021anerf} or more general objects \cite{pumarola2020d,li2020NSFF,Xian2020,park2020nerfies,du2020nerflow,li2021neural}. 
Our method differs from these by tackling general, real-world dynamic scenes from monocular RGB observations and camera parameters only, without using any other auxiliary method to estimate, for example, optical flow or depth.

\section{Method}\label{sec:method} 

Our Non-Rigid Neural Radiance Field (NR-NeRF) approach takes as input a set of $N$ RGB images  $\{\hat{\mathbf{c}}_i\}^{N-1}_{i=0}$ of a non-rigid  scene and their extrinsics  $\{\mathbf{R}_i,\mathbf{t}_i\}^{N-1}_{i=0}$ and  intrinsics $\{\mathbf{K}_i\}^{N-1}_{i=0}$. 
NF-NeRF then finds a single canonical neural radiance volume that can be deformed via ray bending to correctly render each $\hat{\mathbf{c}}_i$. 
Specifically, we collect appearance and geometry information in the static canonical volume $\mathbf{v}$ parametrized by weights~$\theta$.
We model deformations by bending the straight rays sent out by a camera to obtain a deformed rendering of $\mathbf{v}$. 
This ray bending is implemented as a ray bending MLP $\mathbf{b}$ with weights~$\psi$.
It maps, conditioned on the current deformation, 3D points (\textit{e.g.,} sampled from the straight rays) to 3D positions in $\mathbf{v}$. 
The deformation conditioning takes the form of auto-decoded latent codes $\{\mathbf{l}_i\}^{N-1}_{i=0}$ for each image $i$.

\subsection{Background: Neural Radiance  Fields}\label{ssec:NERF} 

We first recap NeRF \cite{mildenhall2020nerf} for rigid scenes. 
NeRF renders a 3D volume into an image by accumulating color, weighted by accumulated transmittance and density, along camera rays. 
The 3D volume is parametrized by an MLP $\mathbf{v}(\mathbf{x},\mathbf{d})=(\mathbf{c}, o)$ that regresses an RGB color $\mathbf{c}=\mathbf{c}(\mathbf{x},\mathbf{d})\in[0,1]^3$ and an opacity $o=o(\mathbf{x})\in[0,1]$ for a point $\mathbf{x}\in\mathbb{R}^3$ on a ray with direction  $\mathbf{d}\in\mathbb{R}^3$.

Consider a pixel $(u,v)$ of an image $\hat{\mathbf{c}}_i$.
For a pinhole camera, the associated ray $\mathbf{r}_{u,v}(j)= \mathbf{o} + j \mathbf{d}(u,v)$ can be calculated using $\mathbf{R}_i,\mathbf{t}_i$, and $\mathbf{K}_i$, which yield the ray origin $\mathbf{o}\in\mathbb{R}^3$ and ray direction $\mathbf{d}(u,v)\in\mathbb{R}^3$.
We can then integrate along the ray from the near plane $j_n$ to the far plane $j_f$ of the camera frustum to obtain the final color $\mathbf{c}$ at $(u,v)$:
\begin{equation}
    \small 
    \mathbf{c}(\mathbf{r}_{u,v})= \int_{j_n}^{j_f} V(j) \cdot  o(\mathbf{r}_{u,v}(j)) \cdot \mathbf{c}(\mathbf{r}_{u,v}(j), \mathbf{d}(u,v)) \,dj \enspace,
\end{equation}
with $V(j) = \exp(-\int_{j_n}^j o(\mathbf{r}_{u,v}(s)) \,ds )$ being the accumulated transmittance along the ray from %
$j_n$ up to $j$. 
In practice, the integrals are approximated by discrete samples~$\mathbf{x}$ along the ray.
NeRF employs a coarse volume $\mathbf{v}_c$ with network weights $\theta_c$ and a fine volume $\mathbf{v}_f$ with network weights $\theta_f$.
Both volumes have the same architecture, but do not share weights: $\theta = \theta_c~\dot{\cup}~ \theta_f$.
When rendering a ray, %
$\mathbf{v}_c$ is accessed first at uniformly distributed samples along the ray. 
These coarse samples are used to estimate the transmittance distribution, from which fine samples are sampled. %
$\mathbf{v}_f$ is then evaluated at the combined set of coarse and fine sample points.
We refer to the original paper for more details.

\noindent\textbf{Adaptations for NR-NeRF.}
We assume Lambertian materials %
and thus remove the view-dependent layers of  rigid NeRF, \textit{i.e.,} we set  $\mathbf{c}=\mathbf{c}(\mathbf{x})$. 
Since each image corresponds to a different deformation of the volume in our non-rigid setting, we also learn a latent code for each time step, which is then used as input for the ray bending network which parameterizes scene deformations.
The weights of this network and the latent codes are shared between $\mathbf{v}_c$ and $\mathbf{v}_f$. 

\subsection{Deformation Model}\label{ssec:deformation_model} 

The original NeRF method~\cite{mildenhall2020nerf} assumes %
rigidity and cannot handle non-rigid scenes.
A na\"ive approach to modeling deformations in the NeRF framework would be to condition the volume on the deformation (\textit{e.g.,} by  conditioning it on time or a deformation  latent code). 
We explore the latter option in the experiments in Sec.~\ref{sec:comparisons}. %
As we will show, apart from not providing hard correspondences, this na\"ive approach only leads to satisfying results when reconstructing the input camera path, but gives implausible results for novel view synthesis.
Instead, we explicitly model the consistency of geometry and appearance across time by disentangling them from the deformation.

We accumulate geometry and appearance from all frames into a single, non-deforming canonical volume.
We employ general space warping (or ray bending)  on top of the static canonical volume to model non-rigid deformations.

For an input image $\hat{\mathbf{c}}_i$ at training time, we want to render the canonical volume such that the image is reproduced.
To that end, we need to un-do the deformation of the specific time step $i$ by mapping the camera rays to the deformation-independent canonical volume.
We first send out straight rays from the input camera%
.
To account for the deformation, we then bend the straight rays such that sampling and subsequently rendering the canonical volume along the bent rays yields $\hat{\mathbf{c}}_i$.
We choose a very unrestricted parametrization of the ray bending, namely an MLP.

Specifically, we implement ray bending as a ray bending network $\mathbf{b}(\mathbf{x}, \mathbf{l}_i)\in\mathbb{R}^3$.
For a point $\mathbf{x}$, for example lying on a straight ray, the network regresses an offset under a deformation represented by $\mathbf{l}_i$.
The offset is then added to $\mathbf{x}$, thus bending the ray.
Finally, we pass the new, bent ray point to the canonical volume, that is: $(\mathbf{c}, o)=\mathbf{v}(\mathbf{x} + \mathbf{b}(\mathbf{x},\mathbf{l}_i))$.
Note that $\mathbf{v}$ is not conditioned on $\mathbf{l}_i$, which leads to the disentanglement of deformation ($\mathbf{b}$ and $\mathbf{l}_i$) from geometry and appearance ($\mathbf{v}$).
We denote the bent version of the straight ray $\bar{\mathbf{r}}$ as $\tilde{\mathbf{r}}_{\mathbf{l}_i}(j) = \bar{\mathbf{r}}(j) + \mathbf{b}(\bar{\mathbf{r}}(j), \mathbf{l}_i) $.

\noindent\textbf{Rigidity Network.}
However, we find that rigid parts of the scene are insufficiently constrained by this formulation. 
We reformulate $\mathbf{b}(\mathbf{x},\mathbf{l}_i)\in\mathbb{R}^3$ as the product of a raw offset $\mathbf{b}'(\mathbf{x},\mathbf{l}_i)$ and a rigidity mask $w(\mathbf{x})\in[0,1]$, \textit{i.e.,}  $\mathbf{b}(\mathbf{x},\mathbf{l}_i)= w(\mathbf{x})\mathbf{b}'(\mathbf{x},\mathbf{l}_i)$.
For rigid objects, we want to prevent deformations and hence desire $w(\mathbf{x})=0$, while for non-rigid objects, we want $w(\mathbf{x})>0$.
This makes it easier for $\mathbf{b}'$ to focus on the non-rigid parts of the scene, which change over time, since rigid parts can get masked out by the rigidity network $w$, which is jointly trained.
Because the rigidity network is not conditioned on the latent code $\mathbf{l}_i$, it is forced to share knowledge about the rigidity of regions in the scene across time steps, which also ensures that parts of the rigid background that can be unregularized at certain time steps are nonetheless reconstructed at all time steps without any deformation.

\subsection{Losses}\label{subs:losses}  
With the architecture specified, we next %
optimize all parameters ($\theta$, $\psi$, $\{\mathbf{l}_i\}_i$) jointly. 
We optimize the network weights as usual but auto-decode the latent codes $\mathbf{l}_i$ \cite{TanMayrovouniotis1995, park2019deepsdf}.

\noindent\textbf{Notation.} 
For ease of presentation, we consider a single time step~$i$ and a single straight ray $\bar{\mathbf{r}}$ with coarse ray points $\bar{C} = \{\bar{\mathbf{r}}(j)\}_{j\in C}$ for a set $C$ of uniformly sampled $j\in[j_n, j_f]$ and fine ray points $\bar{F} = \{\bar{\mathbf{r}}(j)\}_{j\in F}$ for a set $F$ of importance-sampled $j$.
For a latent code $\mathbf{l}$, the bent ray $\tilde{\mathbf{r}}_\mathbf{l}$ gives $\tilde{C} = \{\tilde{\mathbf{r}}_\mathbf{l}(j)\}_{j\in C}$ and $\tilde{F} = \{\tilde{\mathbf{r}}_\mathbf{l}(j)\}_{j\in F}$.
The actual training uses a batch of randomly chosen rays from the training images.

\noindent\textbf{Reconstruction Loss.}
We adapt the data term from NeRF to our non-rigid setting as follows:
\begin{equation}
    L_{\textit{data}} = \lVert \mathbf{c}_c(\tilde{C}) - \hat{\mathbf{c}}(\mathbf{r})\rVert_2^2 + \lVert \mathbf{c}_f(\tilde{C} ~\dot{\cup}~ \tilde{F}) - \hat{\mathbf{c}}(\mathbf{r})\rVert_2^2 \enspace,
\end{equation}
where $\hat{\mathbf{c}}(\mathbf{r})$ is the ground-truth color of the pixel and $\mathbf{c}(S)$ is the estimated ray color on the set $S$ of discrete ray points.

While this reconstruction loss yields satisfactory results 
along the space-time camera trajectory of the input recording, we show later in Sec.~\ref{sec:ablation} that it leads to undesirable renderings for novel views.
We thus find it necessary to regularize the bending of rays with further priors.

\noindent\textbf{Offsets Loss.}
We regularize the offsets with a loss on their magnitude.
Since we want visually unoccupied space (\textit{i.e.,} air) to be compressible and not hinder the optimization, we weigh the loss at each point by its opacity.
However, this would still apply a high weight to completely occluded points along the ray, which leads to artifacts when rendering novel views.
We thus additionally weigh by transmittance:
\begin{equation}
 L_\textit{na\"ive offsets} = \frac{1}{\lvert C \rvert} \sum_{j\in C} \alpha_j \cdot  \lVert \mathbf{b}(\bar{\mathbf{r}}(j), \mathbf{l}) \rVert^{2-w(\tilde{\mathbf{r}}(j))}_2,
\end{equation}
where we weigh each point by transmittance and occupancy $\alpha_j=V(j) \cdot o(\tilde{\mathbf{r}}(j))$.
We do not back-propagate into $\alpha_j$.

However, as we show in 
Sec.~\ref{sec:results}, 
we find that applying the offsets loss to the masked offsets leads to an unstable background in novel  views. 
We hypothesize that this is due to the multiplicative ambiguity between unmasked offsets and rigidity mask.
We find that applying the loss to the regressed rigidity mask and raw offsets separately works better:
\begin{equation}
\small 
    L_\textit{offsets} = \frac{1}{\lvert C \rvert} \sum_{j\in C} \alpha_j \cdot \big( \lVert \mathbf{b}'(\bar{\mathbf{r}}(j), \mathbf{l}) \rVert^{2-w(\tilde{\mathbf{r}}(j))}_2 + \omega_\text{rigidity}w(\bar{\mathbf{r}}(j))\big) ,
\end{equation}
where we penalize $\mathbf{b}'$ instead of $\mathbf{b}$.
The exponent of the first term is a tweak to get two desirable properties that neither an $\ell_1$ nor an $\ell_2$ loss fulfills: %
For non-rigid objects ($w$ closer to $1$), it becomes an $\ell_1$ loss, which has two advantages: (1)~the gradient is independent of the magnitude of the offset, so unlike with an $\ell_2$ loss, small and large offsets/motions are treated equally, and (2)~relative to an $\ell_2$ loss, it encourages sparsity in the offsets field, which fits our scenes.
For rigid objects ($w$ closer to $0$), it becomes an $\ell_2$ loss, which tampers off in its gradient magnitude as the offset magnitude approaches 0, preventing noisy gradients that an $\ell_1$ loss has for the tiny offsets of rigid objects.

\noindent\textbf{Divergence Loss.} 
Since the offsets loss only constrains visible areas, we introduce additional regularization of hidden areas.
Inspired by local, isometric shape preservation from computer graphics, like as-rigid-as-possible regularization for surfaces \cite{sorkine2007rigid,igarashi2005rigid} or volume preservation for volumes \cite{Slavcheva2017}, we seek to preserve the local shape after deformation.
To that end, we propose to regularize the absolute value of the divergence of the offsets field. 
The Helmholtz decomposition \cite{bhatia2012helmholtz} allows to split any twice-differentiable 3D vector field on a bounded domain into a sum of a rotation-free %
and a divergence-free vector fields. 
Thus, by penalizing the divergence, we encourage the vector field to be composed  primarily of translations and rotations, effectively preserving volume. 
The divergence loss is:
\begin{equation}
    L_\textit{divergence} =  \frac{1}{\lvert C \rvert} \sum_{j\in C} w'_j \cdot \lvert \operatorname{div} (\mathbf{b}(\bar{\mathbf{r}}(j), \mathbf{l})) \rvert^2 \enspace,
\end{equation}
where we do not back-propagate into $w'_j = o(\tilde{\mathbf{r}}(j))$, and we take the divergence $\operatorname{div}$ of $\mathbf{b}$ w.r.t. the position $\bar{\mathbf{r}}(j)$.

We employ FFJORD's~\cite{grathwohl2018ffjord} fast, unbiased divergence estimation, which is three times less computationally expensive than an exact computation. 
The divergence is defined as:
\begin{equation}
    \operatorname{div}(\mathbf{b}(\mathbf{x})) = \operatorname{Tr}\bigg(\frac{\text{d}\mathbf{b}(\mathbf{x})}{\text{d}\mathbf{x}}\bigg) = \frac{\partial \mathbf{b}(\mathbf{x})_x}{\partial x} + \frac{\partial \mathbf{b}(\mathbf{x})_y}{\partial y} + \frac{\partial \mathbf{b}(\mathbf{x})_z}{\partial z} \enspace,
\end{equation}
where $\mathbf{b}(\mathbf{x})_k \in \mathbb{R}$ is the $k$-th component of $\mathbf{b}(\mathbf{x})$, $\operatorname{Tr}(\cdot)$ is the trace operator, and  $\frac{\text{d}\mathbf{b}(\mathbf{x})}{\text{d}\mathbf{x}}$ is the $3\times 3$ Jacobian matrix.
Na\"ively computing the divergence with PyTorch's automatic differentiation requires three backward passes, one for each term of the sum.
Instead, the authors of FFJORD~\cite{grathwohl2018ffjord} use Hutchinson's trace estimator~\cite{hutchinson1989stochastic}:
\begin{equation}\label{eq:monte_carlo}
    \operatorname{Tr}(\mathbf{A}) = \mathbb{E}_\mathbf{e}[\mathbf{e}^T\mathbf{A}\mathbf{e}] \enspace.
\end{equation}
Here, $\mathbf{e}$ is Gaussian-distributed.
The single-sample Monte-Carlo estimator implied by this expectation can be computed with a single backward pass.

\noindent\textbf{Full Loss.}
We combine all losses to obtain the full loss:
\begin{equation}
    L = L_\text{data} + \omega_\text{offsets} L_\text{offsets} + \omega_\text{divergence} L_\text{divergence} \enspace,
\end{equation}
where the weights $\omega_\text{rigidity}$, $\omega_\text{offsets}$, and $\omega_\text{offsets}$ are scene-specific since we consider a variety of non-rigid scene types. 
Our implementation uses the structure-from-motion method COLMAP~\cite{schonberger2016structure} to estimate the camera parameters. 
For further training and implementation details, we refer to the supplemental material. 
We will release our source code.

\section{Results}\label{sec:results} 
We present qualitative results of our method, including rigidity scores and correspondences, by rendering into input and novel spatio-temporal views in Sec.~\ref{sec:qualitative}.%
Turning %
to the inner workings, Sec.~\ref{sec:ablation} investigates the crucial design choices we made to improve novel view quality.
We conclude the evaluation of our approach in Sec.~\ref{sec:comparisons} by comparing to prior work and a baseline approach.
Finally, we show simple scene-editing results in Sec.~\ref{sec:editing}. 
In the supplemental material, we provide information on data capture and show extensions to multi-view data and view-dependent effects. %

\begin{figure}

\setlength{\tabcolsep}{0.em} %
\def\arraystretch{0.} %

\begin{tabular}{ccc}

  \includegraphics[width=.333\columnwidth]{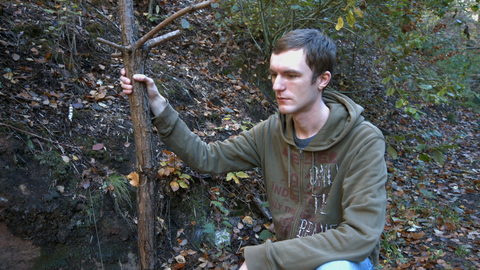}
  &
  \includegraphics[width=.333\columnwidth]{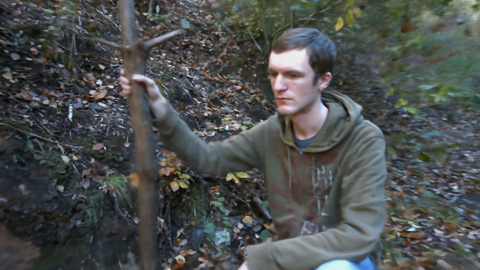}
  &
  \includegraphics[width=.333\columnwidth]{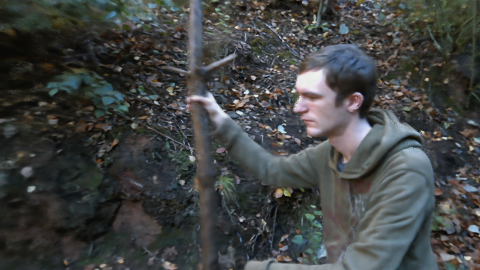}
  \\

  \includegraphics[width=.333\columnwidth,trim={0 50 0 50},clip]{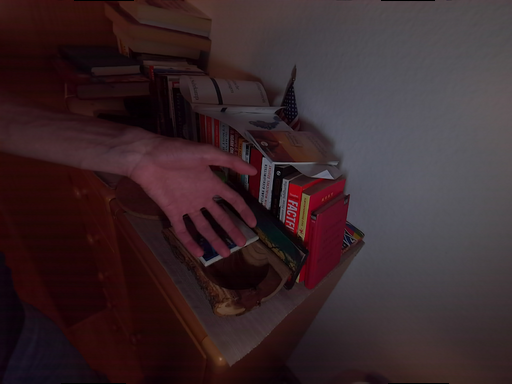}
  &
  \includegraphics[width=.333\columnwidth,trim={0 50 0 50},clip]{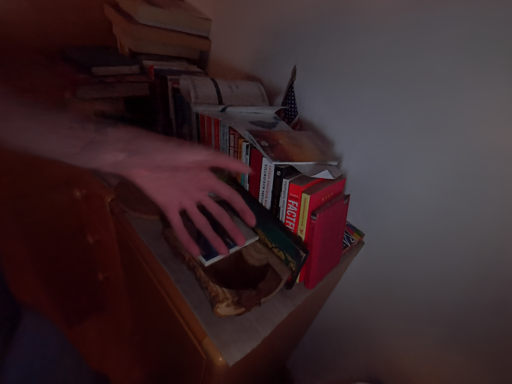}
  &
  \includegraphics[width=.333\columnwidth,trim={0 50 0 50},clip]{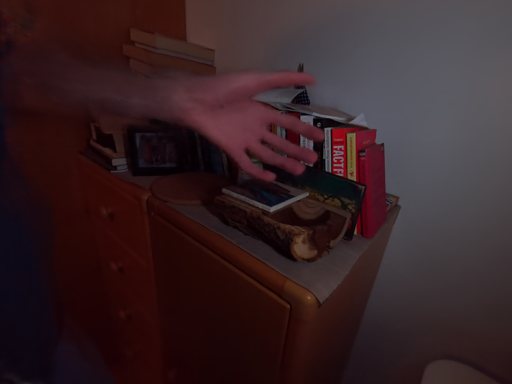}
  \\

\end{tabular}
  \caption{The input (left) is reconstructed by NR-NeRF (middle) and rendered into a novel view (right). }
  \label{fig:novel_view}
\end{figure}

\begin{figure}
\centering
\setlength{\tabcolsep}{0.05em} %
\def\arraystretch{0.1} %
\resizebox{\columnwidth}{!}{%
\begin{tabular}{cccc|cccc}
  \rotatebox[origin=c]{90}{Input}
  &
  \raisebox{-0.5\height}{\includegraphics[height=.15\textheight,trim={100 0 100 0},clip]{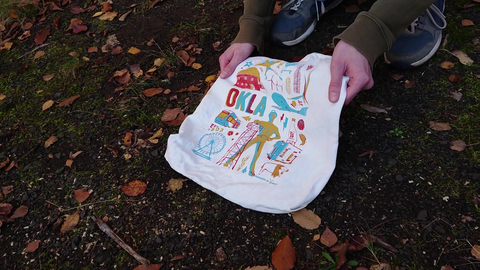}}
  &

  \raisebox{-0.5\height}{\includegraphics[height=.15\textheight]{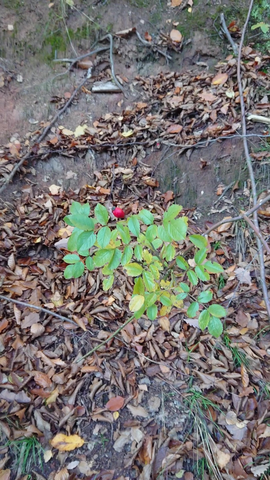}}

  &
  \raisebox{-0.5\height}{\includegraphics[height=.15\textheight]{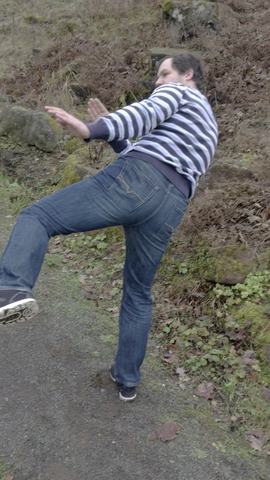}}
  &
  
  \rotatebox[origin=c]{90}{Color}
  &
  \raisebox{-0.5\height}{\includegraphics[height=.15\textheight,trim={100 0 100 0},clip]{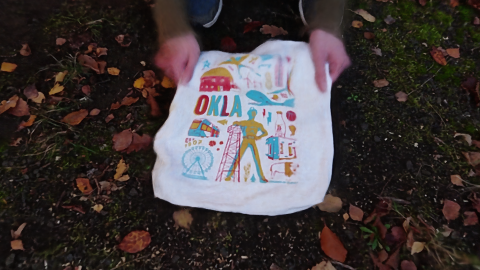}}

  &
  \raisebox{-0.5\height}{\includegraphics[height=.15\textheight]{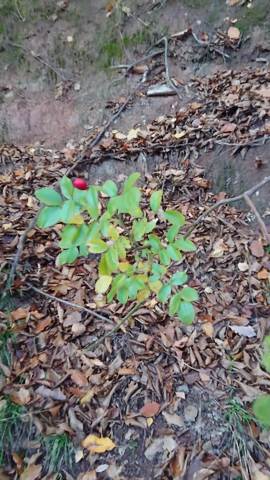}}

  &
  \raisebox{-0.5\height}{\includegraphics[height=.15\textheight]{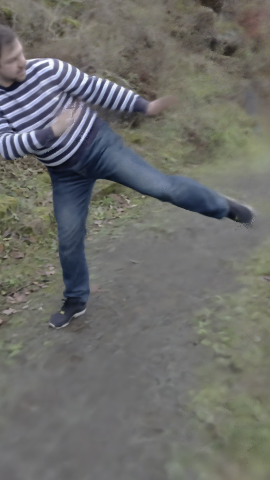}}
  \\
  
  \rotatebox[origin=c]{90}{Rigidity}
  &
  \raisebox{-0.5\height}{\includegraphics[height=.15\textheight,trim={100 0 100 0},clip]{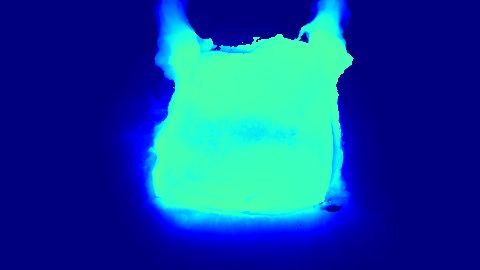}}

  &
  \raisebox{-0.5\height}{\includegraphics[height=.15\textheight]{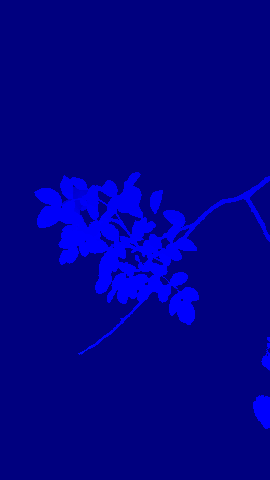}}

  &
  \raisebox{-0.5\height}{\includegraphics[height=.15\textheight]{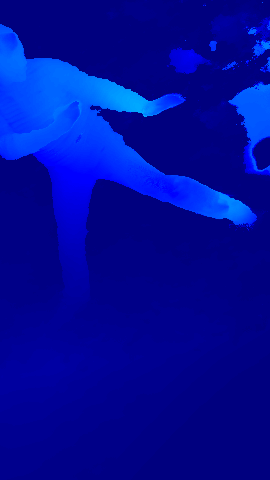}}
  &

  \rotatebox[origin=c]{90}{Correspondences}
  &
  \raisebox{-0.5\height}{\includegraphics[height=.15\textheight,trim={100 0 100 0},clip]{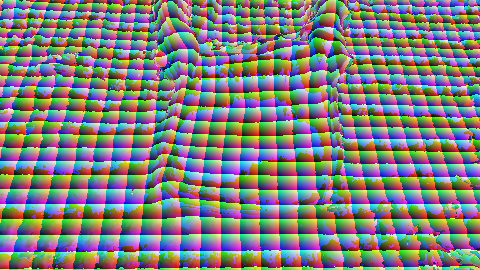}}

  &
  \raisebox{-0.5\height}{\includegraphics[height=.15\textheight]{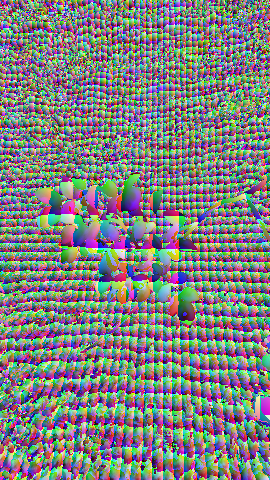}}

  &
  \raisebox{-0.5\height}{\includegraphics[height=.15\textheight]{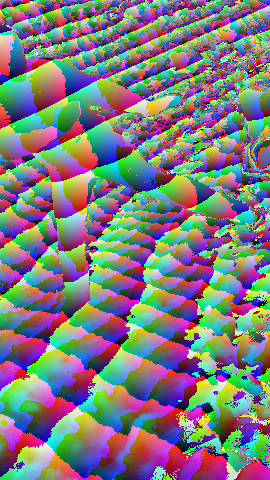}}
  \\

\end{tabular}
}
  \caption{NR-NeRF can render a deformed state captured at a certain time into a novel view. We visualize here this novel-view rendering and additional modalities as seen from the novel view, namely rigidity scores and correspondences%
  .  } 
  \label{fig:modalities}
\end{figure}

\subsection{Qualitative Results}\label{sec:qualitative}
We present qualitative results of NR-NeRF by rendering the scene from input and novel spatio-temporal views.
We also visualize the additional outputs of our method. %

\noindent\textbf{Input Reconstruction and Novel View Synthesis.}
Fig.~\ref{fig:novel_view} shows examples of input reconstruction and novel view synthesis with NR-NeRF. %
As the center column shows, the input is reconstructed faithfully.
This enables high-quality novel view synthesis,  example results can be found in the third column.
We can freely move the camera in areas around the original camera paths and specify the time 
step. 

\noindent\textbf{Rigidity.}
NR-NeRF estimates rigidity scores without supervision to improve background stability in novel-view renderings.
We show examples in Fig.~\ref{fig:modalities} and find that the background is consistently scored as highly rigid while foreground is correctly estimated to be rather non-rigid. 

\noindent\textbf{Correspondences.}
Another side effect of our proposed approach is the ability to estimate consistent dense 3D correspondences into the canonical model across different camera views and time steps.
Fig.~\ref{fig:modalities} shows examples.

\subsection{Ablation Study}\label{sec:ablation}

After this look at the outputs of NR-NeRF, we now turn to its internal workings.
Specifically, since we aim for convincing novel-view renderings, we take a closer look at the impact of some of our design choices on the foreground and background stability of our novel-view results.

\noindent\textbf{Setup.} 
We investigate the necessity of all regularization losses by removing each loss individually and all of them at once.
Next, we remove the rigidity network to see the impact on background stability.
Finally, we determine whether applying the offsets loss separately on both the regressed rigidity and the unmasked offsets, \textit{i.e.,} $L_\textit{offsets}$ in our method, or directly on the masked offsets, $L_\textit{na\"ive~offsets}$, works better.

\begin{figure}
\centering
\setlength{\tabcolsep}{0.2em} %
\def\arraystretch{0.3} %
\begin{tabular}{ccc}
  \includegraphics[width=.15\textwidth,trim={120 0 50 0},clip]{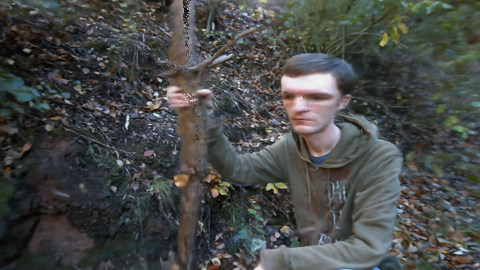}
  &
  \includegraphics[width=.15\textwidth,trim={120 0 50 0},clip]{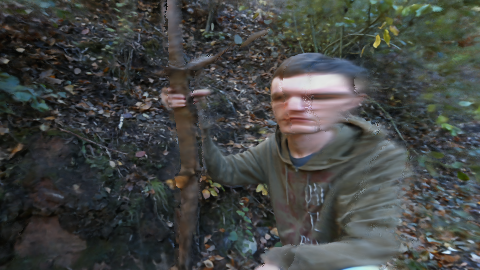}
  &
  \includegraphics[width=.15\textwidth,trim={120 0 50 0},clip]{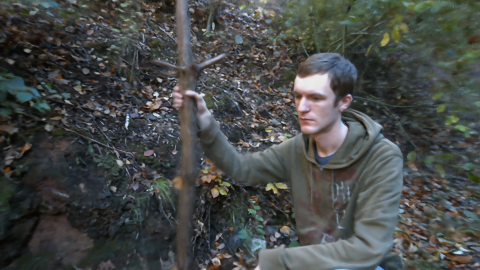}
  \\

  \small Without $L_\mathit{divergence}$ &  \small No regularization &  \small Ours
  
\end{tabular}
  \caption{Ablation Study. We render the scene into novel views to determine the stability of the non-rigid part after removing the divergence loss, all regularization losses, and none of the losses. 
  }
  \label{fig:divergence_qualitative}
\end{figure}

\begin{figure}

\setlength{\tabcolsep}{0em} %
\def\arraystretch{0.3} %
\begin{tabular}{cccc}
\multicolumn{1}{l}{\large{a)}} & & & \\
\multicolumn{2}{c}{\includegraphics[width=.45\columnwidth]{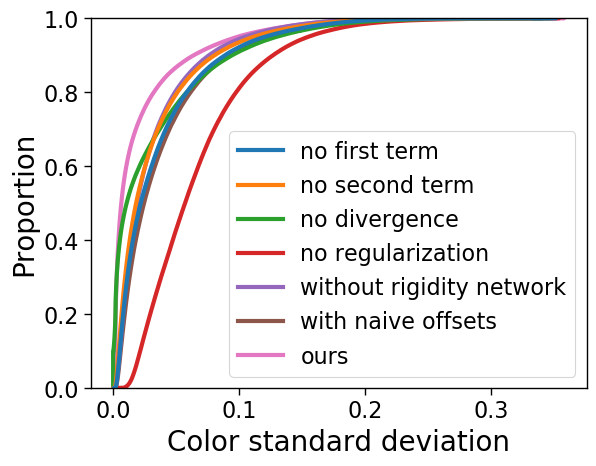}}~~
&
\multicolumn{2}{c}{\includegraphics[width=.45\columnwidth]{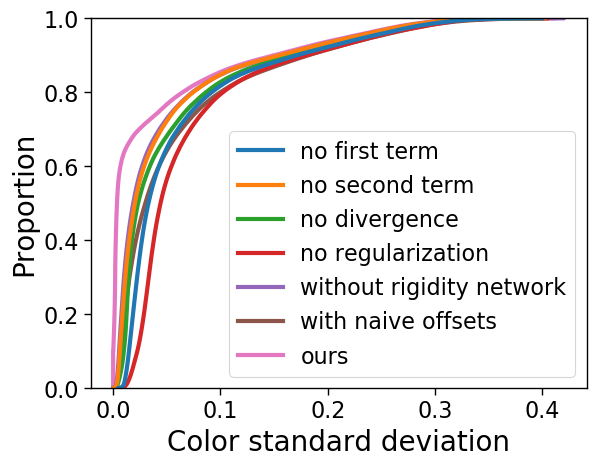}}
\\

\multicolumn{2}{c}{Left scene}~~
&
\multicolumn{2}{c}{Right scene}
\\

\multicolumn{1}{l}{\large{b)}} & & & 
\\

\multicolumn{2}{c}{\includegraphics[width=.44\columnwidth]{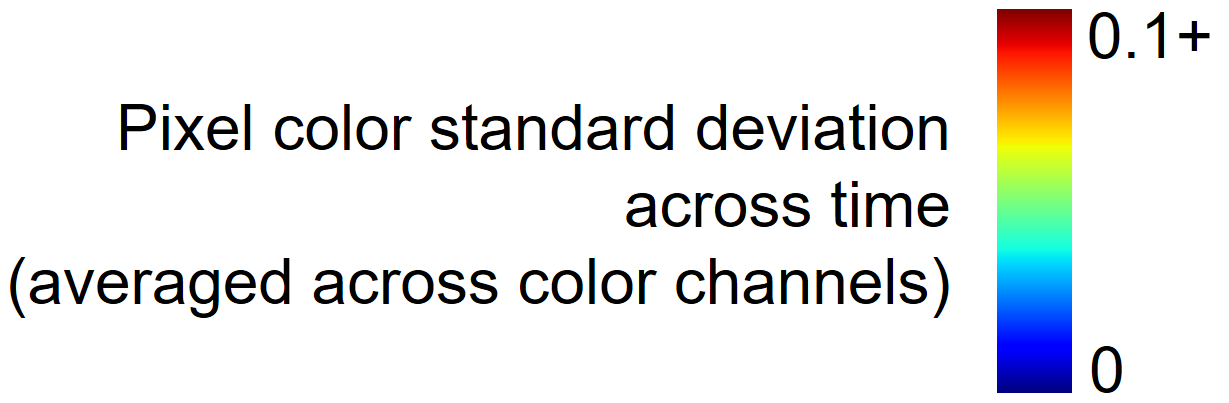}}
   &
  \includegraphics[width=.25\columnwidth]{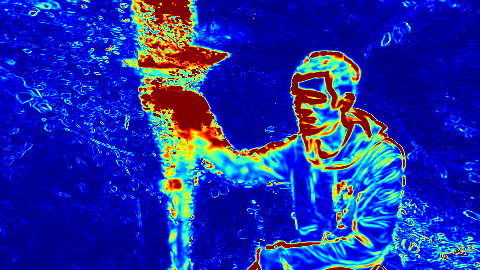}
  &
   \includegraphics[width=.25\columnwidth]{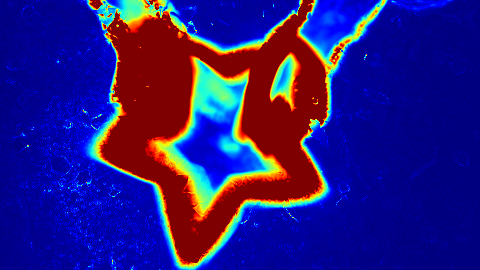}~~
  \\
     &  &
  \multicolumn{2}{c}{\footnotesize Ours}
   \\
\tiny\color{white} . & & & 
\\

\includegraphics[width=.25\columnwidth]{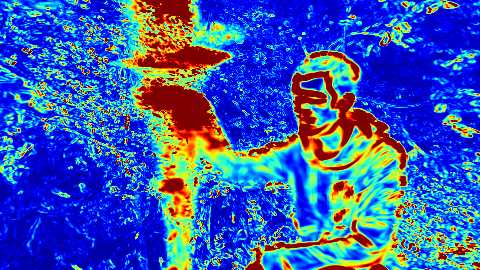}
&
\includegraphics[width=.25\columnwidth]{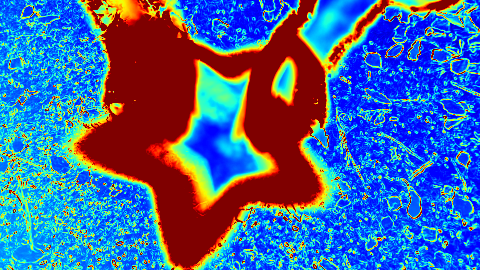}~~
  &
  \includegraphics[width=.25\columnwidth]{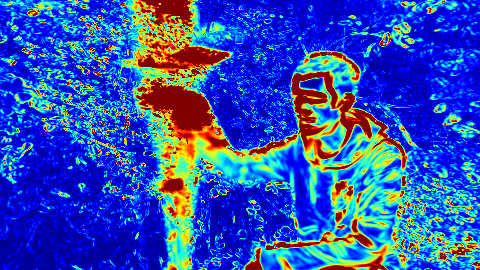}
  &
  \includegraphics[width=.25\columnwidth]{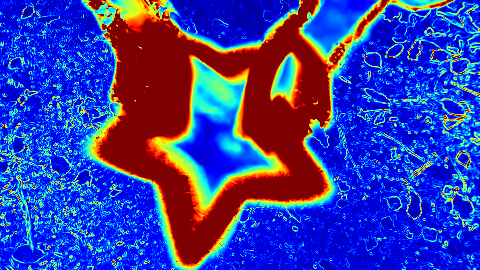}~~
  \\
  \multicolumn{2}{c}{\footnotesize No first term of offsets loss $L_\textit{offsets}$}
  &
    \multicolumn{2}{c}{\footnotesize No second term of offsets loss $L_\textit{offsets}$}
  \\
  \tiny\color{white} . & & & \\

  \includegraphics[width=.25\columnwidth]{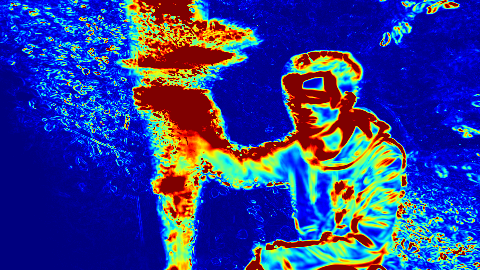}
  &
  \includegraphics[width=.25\columnwidth]{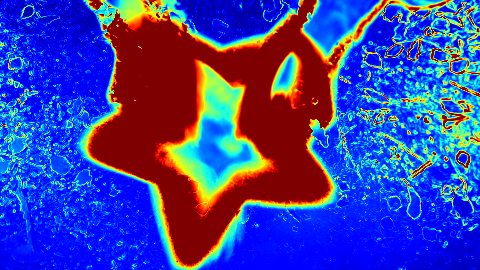}~~
  &
  \includegraphics[width=.25\columnwidth]{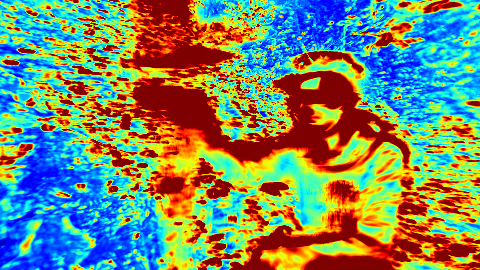}
  &
  \includegraphics[width=.25\columnwidth]{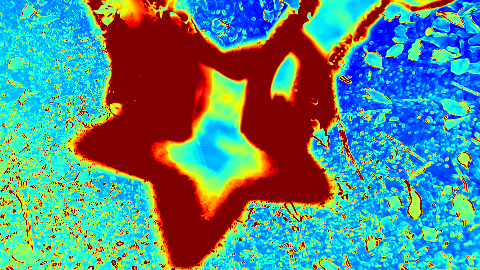}~~
  \\
  \multicolumn{2}{c}{\footnotesize Without divergence loss $L_\textit{divergence}$}
  &
    \multicolumn{2}{c}{\footnotesize Without any regularization loss}
  \\
  \tiny\color{white} . & & & \\

  \includegraphics[width=.25\columnwidth]{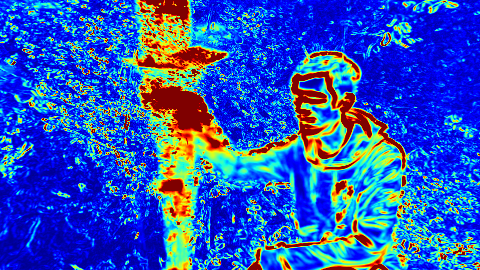}
  &
  \includegraphics[width=.25\columnwidth]{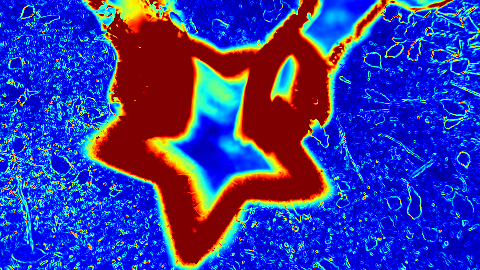}~~
  &
  \includegraphics[width=.25\columnwidth]{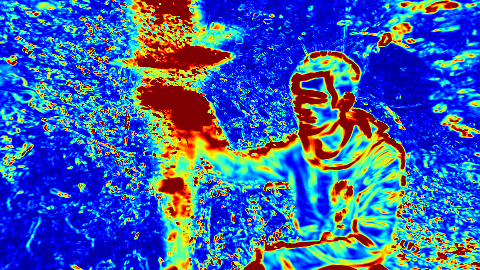}
  &
  \includegraphics[width=.25\columnwidth]{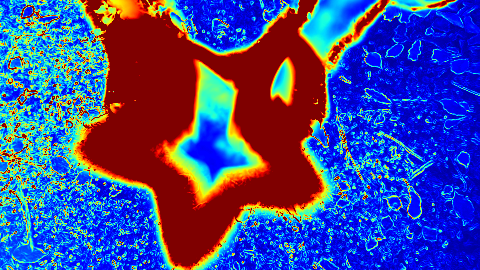}~~
  \\
  \multicolumn{2}{c}{\footnotesize Without rigidity network $w$}
  &
    \multicolumn{2}{c}{\footnotesize With $L_\textit{na\"ive offsets}$}
  \\
  \tiny\color{white} . & & &\\

\end{tabular}%
  \caption{Ablation Study. We quantify the impact of our main design choices on background stability. To that end, we render the entire input sequence into a fixed novel view \emph{for all time steps} and compute the standard deviation of each pixel's color across time to measure color changes and hence background stability. \textbf{a)}~We show cumulative plots across all pixels, where our full method (left-most curve) has the most stable background. \textbf{b)}~We then show how those instabilities are distributed in the scene. The results of NR-NeRF show the least instability in the background.%
  }
  \label{fig:background_instability}
\end{figure}

\noindent\textbf{Results.}
We find that %
$L_\textit{divergence}$ 
is crucial for stable deformations of the non-rigid objects in the foreground, see Fig.~\ref{fig:divergence_qualitative}. %
On the other side, the interplay of \emph{all} of the remaining design choices is necessary to stabilize the rigid background as Fig.~\ref{fig:background_instability} shows.
The supplemental video contains video examples that highlight the instability in these cases.

\subsection{Comparisons}\label{sec:comparisons}

Having only considered our method in isolation so far, we next compare NR-NeRF to prior work and a baseline. 
In this section, we split the images into training and test sets by partitioning the temporally-ordered images into consecutive blocks of length 16 each, with the first twelve for the training set and the remaining four for the test set.

\noindent\textbf{Prior Work and Baseline.}
We start with the trivial baseline of rigid NeRF~\cite{mildenhall2020nerf}, which cannot handle dynamic scenes.
We consider two variants: view-dependent rigid NeRF, as in the original method~\cite{mildenhall2020nerf}, and view-independent rigid NeRF, where we remove the view-direction conditioning.
We next introduce \emph{na\"ive NR-NeRF}, which  adds na\"ive support for dynamic scenes to rigid  NeRF: We condition the neural radiance fields  volume on the latent code~$\mathbf{l}_i$,  \textit{i.e.,} $(\mathbf{c},  o)=\mathbf{v}(\mathbf{x}, \mathbf{l}_i)$. 
For test images $i$, we backpropagate gradients into the corresponding latent code $\mathbf{l}_i$.
We do the same for NR-NeRF in order to optimize for the test latent codes. 
Note that test images \emph{solely} influence test latent codes, as is typical for auto-decoding~\cite{park2019deepsdf}.
Finally, we compare to Neural Volumes~\cite{lombardi2019neural}, for which we use the official code release.
We consider two variants: (1) as in \cite{lombardi2019neural}, the geometry and appearance template is conditioned on the latent code (\textit{NV}), and (2) the geometry and appearance template are independent of the latent code (\textit{modified NV}).

\begin{figure*}

\setlength{\tabcolsep}{0.05em} %
\def\arraystretch{0.1} %
\centering
\begin{tabular}{cccccccc}

  &
  \small Input
  &
  \small Ours
  &
  \small Na\"ive NR-NeRF
  &
  \small Rigid NeRF
  &
  \small Rigid NeRF
  &
  \small Neural Volumes
  &
  \small Neural Volumes
  \\
  
  &
  &
  &
  &
  \small (view-dep.)
  &
  \small (not view-dep.)
  &

  &
  \small (modified)
  \\

  &
  \includegraphics[width=.12\textwidth]{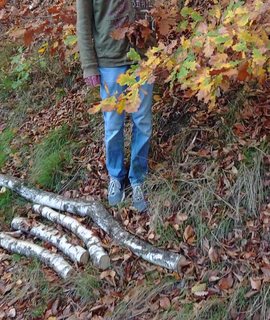}
  &
  \includegraphics[width=.12\textwidth]{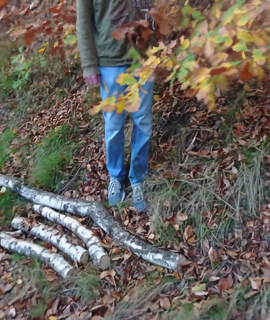}
  &
  \includegraphics[width=.12\textwidth]{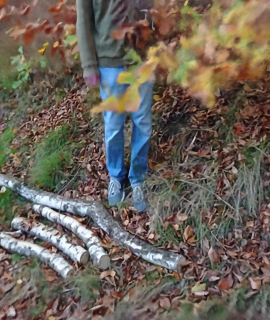}
  &
  \includegraphics[width=.12\textwidth]{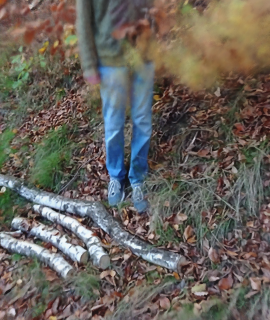}
  &
  \includegraphics[width=.12\textwidth]{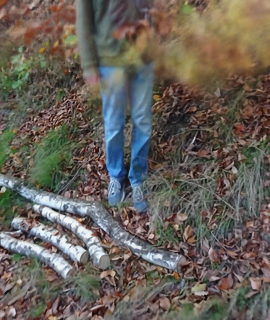}
  &
  \includegraphics[width=.12\textwidth]{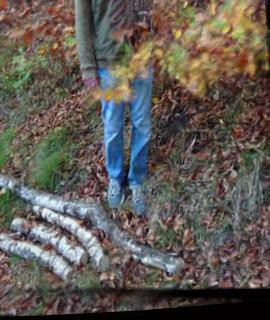}
  &
  \includegraphics[width=.12\textwidth]{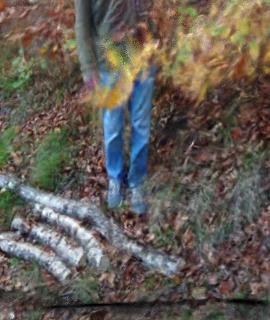}
  \\
  
  &
  
  &
  \includegraphics[width=.12\textwidth]{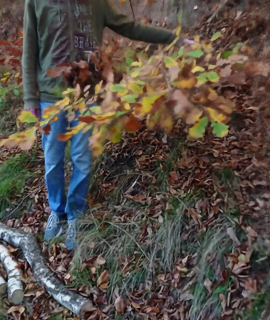}
  &
  \includegraphics[width=.12\textwidth]{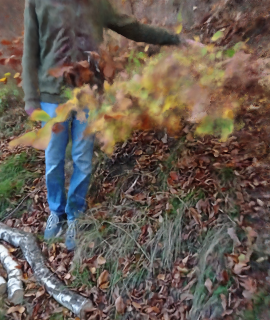}
  &
  \includegraphics[width=.12\textwidth]{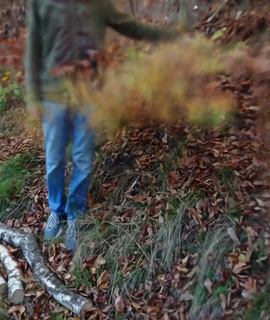}
  &
  \includegraphics[width=.12\textwidth]{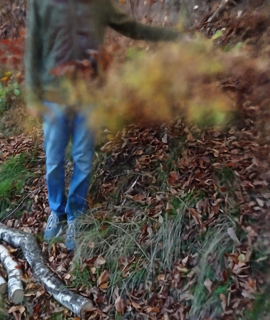}
  &
  \includegraphics[width=.12\textwidth]{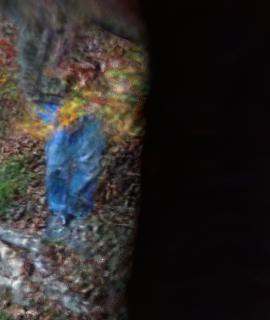}
  &
  \includegraphics[width=.12\textwidth]{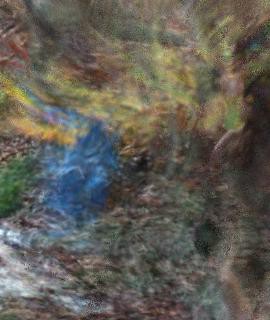}
  \\

\end{tabular}
  \caption{We compare input reconstruction quality (first row) and novel view synthesis quality (second row). Only our method synthesizes sharp novel views. 
  } 
  \label{fig:qualitative_comparison}
\end{figure*}

\noindent\textbf{Input Reconstruction.}
We first consider input reconstruction quality on the training set to verify the plausibility of the learned representations.
See Fig.~\ref{fig:qualitative_comparison}. %
We find that na\"ive NR-NeRF and both variants of Neural Volumes perform very well on this task, similar to our method.
However, rigid NeRF's not accounting for deformations leads to blur.

\noindent\textbf{Novel View Synthesis.}
We next evaluate novel-view performance qualitatively and quantitatively on the test sets. %
Fig.~\ref{fig:qualitative_comparison} %
contains novel-view results of all methods.
Both versions of Neural Volumes give implausible results that are in some cases only barely recognizable.
The two rigid NeRF variants show blurry, static results similar to the training reconstruction results earlier.
While the still images in Fig.~\ref{fig:qualitative_comparison} show some undesirable artifacts like blurrier or less stable results compared to ours, we refer to the supplemental video to see na\"ive NeRF's temporal inconsistencies
, especially on spatio-temporal trajectories different from the input.

After the qualitative overview, we now evaluate the novel-view results of the methods considered quantitatively.
We use the same three metrics as NeRF~\cite{mildenhall2020nerf}.
We use PSNR and SSIM~\cite{ssim2004} as conventional metrics for image similarity, where higher is better.
In addition, we use a learned perceptual metric, LPIPS~\cite{zhang2018perceptual}, where lower is better.
Tab.~\ref{tab:quantitative} contains the quantitative results. %
Our method obtains the best SSIM and LPIPS scores, and the second-best PSNR after na\"ive NR-NeRF.
As we saw in the input reconstruction results, na\"ive NR-NeRF is competitive for settings that are close to the input spatio-temporal trajectory, as is the case for our test sets.
We, therefore, next evaluate more challenging novel view scenarios with a spatio-temporal trajectory significantly different from the input. 
Since we do not have access to ground-truth novel view data, we focus on background stability for a spatially fixed camera.

\begin{table}[]
\centering
\resizebox{0.85\columnwidth}{!}{%
\begin{tabular}{rc|c|c|c|c|c|c|}
& &  Ours &  Na\"ive  &  Rigid &  Rigid & NV & NV\\
& &  &  & \small (cond.) & \small (no cond.) & & \small (mod.)\\
\cline{1-8} 
PSNR & $\uparrow$ & $\mathit{24.70}$ & $\mathbf{25.83}$ & 22.24 & 21.88 & 14.13 & 14.10\\
SSIM & $\uparrow$ & $\mathbf{0.758}$ & $\mathit{0.738}$ & 0.662 & 0.659 & 0.259 & 0.263 \\
LPIPS & $\downarrow$ & $\mathbf{0.197}$ & $\mathit{0.226}$ & 0.309 & 0.313 & 0.580 & 0.583 \\
\hline
\end{tabular}%
}
\caption{Quantitative Results Averaged Across Scenes. We evaluate our method%
, na\"ive NR-NeRF, rigid NeRF~\cite{mildenhall2020nerf} (1) with view conditioning and (2) without view conditioning, and Neural Volumes~\cite{lombardi2019neural} (1) without and (2) with modifications. For PSNR and SSIM \cite{ssim2004}, higher is better. For LPIPS \cite{zhang2018perceptual}, lower is better.}
\label{tab:quantitative}
\end{table}

\begin{figure}

\setlength{\tabcolsep}{0.em} %
\def\arraystretch{0.7} %
\begin{tabular}{cccc}

  \includegraphics[width=.28\columnwidth]{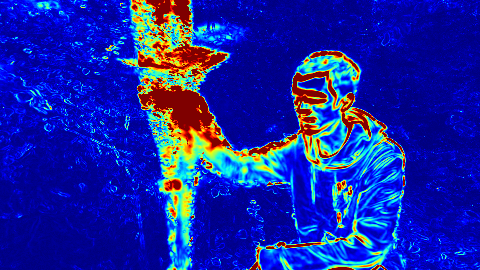}
  &
  \includegraphics[width=.21\columnwidth]{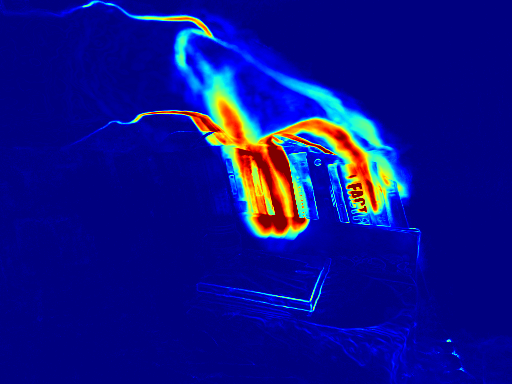}~~
  &
  \includegraphics[width=.28\columnwidth]{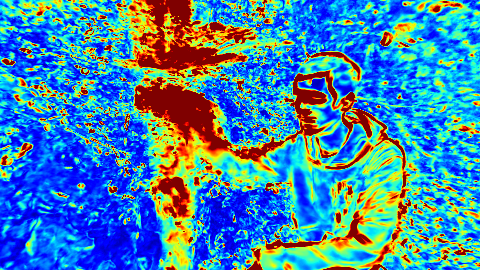}
  &
  \includegraphics[width=.21\columnwidth]{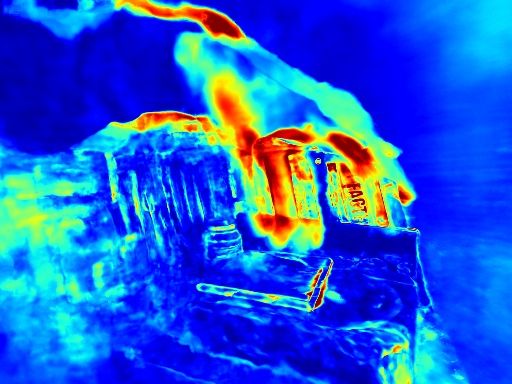}
  \\  
  \multicolumn{2}{c}{\small Ours}
  &
        \multicolumn{2}{c}{\small Na\"ive NR-NeRF}
  \\

  \includegraphics[width=.28\columnwidth]{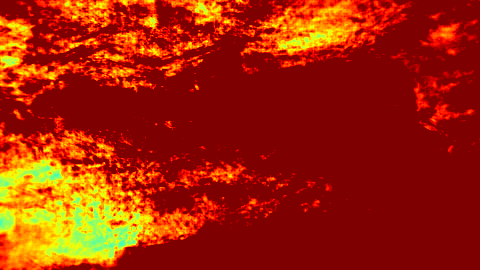}
  &
  \includegraphics[width=.21\columnwidth]{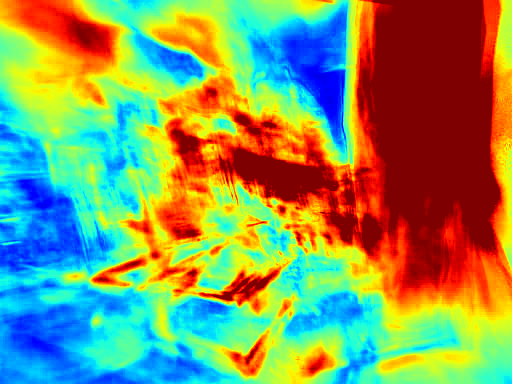}~~
  &
    \includegraphics[width=.28\columnwidth]{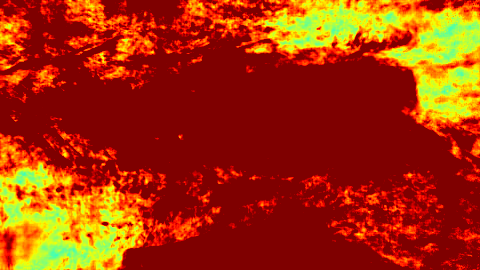}
  &
  \includegraphics[width=.21\columnwidth]{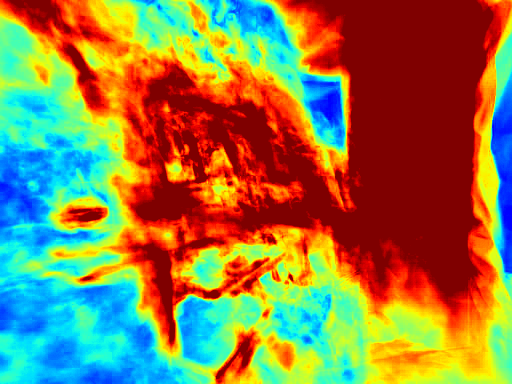}
  \\
  \multicolumn{2}{c}{\small Neural Volumes }
  &
    \multicolumn{2}{c}{\small Neural Volumes (modified)}

\end{tabular}%

  \caption{We compare background stability. See Fig.~\ref{fig:background_instability} for an explanation. We use all test time steps here. %
  The results of NR-NeRF show the least instability.
  }
  \label{fig:comparison_background_instability}
\end{figure}

\noindent\textbf{Background Stability.}
While a moving camera during rendering can obfuscate background instability, we found that stabilizing the background for fixed novel-view renderings matters for perceptual fidelity but is  difficult to achieve.
We thus quantitatively evaluate on this challenging task here.
In Fig.~\ref{fig:comparison_background_instability}, we compare the background stability of our method, na\"ive NR-NeRF, and the Neural Volumes variants.
We exclude rigid NeRF since it is static by design.
We find that our method leads to significantly more stable background synthesis than the other methods, and we refer to the supplemental material for further results. %

\subsection{Simple Scene Editing}\label{sec:editing}

We can manipulate the learned model in several simple ways: foreground removal, temporal super-sampling, deformation exaggeration and dampening, and forced background stabilization. 
We discuss foreground removal here and the other editing tasks in the supplemental material due to space constraints.

\begin{figure}
    \centering
  
  \includegraphics[width=.2\textwidth]{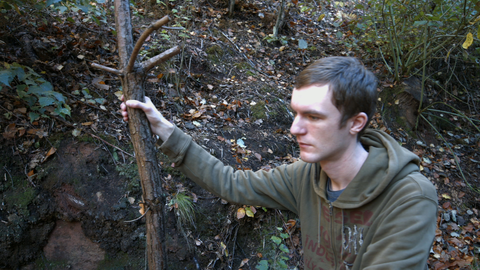}%
  \hspace{0.2em}
  \includegraphics[width=.2\textwidth]{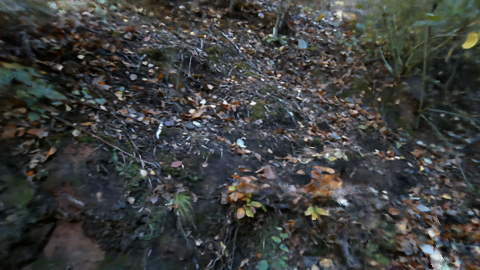} 
  \caption{(Left) the ground-truth input image and (right) a rendering without non-rigid foreground. %
  }
  \label{fig:foreground_removal}
\end{figure}

Our representation enables us to remove a potentially occluding non-rigid object from the foreground, leaving only the unoccluded background.
Assuming the rigidity network assigns higher scores to non-rigid objects than to rigid (background) objects, we can threshold them at test time to segment the canonical volume into rigid and non-rigid parts.
We can then set the non-rigid part transparent,
see Fig.~\ref{fig:foreground_removal}. %

\section{Limitations} 
\label{sec:limitations} 
For simplicity, the discrete integration along the bent ray uses the interval lengths given by the straight ray.
As we build on NeRF, our method is similarly slow.
All else being equal, ray bending increases runtime by about 20\%. 
However, due to fewer rays and points sampled, we train for 6 hours. 
We can thus train multiple NR-NeRFs (to find appropriate loss weights) in a time similar to other NeRF-based methods \cite{mildenhall2020nerf}. 
Neural Sparse Voxel Fields \cite{liu2020neural} are a promising direction to speed up NeRF-like methods. 
The background needs to be fairly close to the foreground, an issue we ``inherit'' from NeRF and which could be addressed similarly to NeRF++ \cite{zhang2020nerfpp}. 
Since we use a deformation model that does not go from the canonical space to the deformed space, we cannot obtain exact correspondences between images captured at different time steps, but instead need to use a nearest neighbor approximation.
We do not account for appearance changes that are due to deformation or lighting changes.
For example, temporally changing shadowing in the input images is an issue. %
Foreground removal can fail if a part of the foreground is entirely static. %
Rendering parts of the scene barely or not at all observed in the training data would not lead to realistic results. 
Motion blur in input images is not modeled and would lead to artifacts.
The background needs to be static and dominant enough for structure-from-motion \cite{schonberger2016structure} to estimate correct extrinsics. 
Since our problem is severely under-constrained, we employ strong regularization, which leads to a trade-off between sharpness and stability on some scenes%
. 

\section{Conclusion}\label{sec:conclusions} 
We presented a method for free viewpoint rendering of a dynamic scene using just a monocular video as input.
Several high-quality reconstruction and novel view synthesis results of general dynamic scenes, as well as unsupervised, yet plausible rigidity scores and dense 3D correspondences demonstrate the capabilities of the proposed method.
Our results suggest that space warping in the form of ray bending is a promising deformation model for volumetric representations like NeRF.
Furthermore, we have demonstrated that background instability, a problem also noted by concurrent work \cite{park2020nerfies}, can be %
mitigated in an unsupervised fashion by learning a rigidity mask%
. 
The extensions to multi-view data and view dependence invite future work on more constrained settings for higher quality.
Although rather rudimentary, we have shown that NR-NeRF enables several scene-editing tasks, and we look forward to further work in the direction of editable neural representations.

\noindent\textbf{Acknowledgements.} 
All data capture and evaluation was done at MPII and Volucap. 
We thank Volucap for providing the multi-view data. 
Research conducted by Ayush Tewari, Vladislav Golyanik and Christian Theobalt at MPII was supported in part by the ERC Consolidator Grant 4DReply (770784). 
This work was also supported by a Facebook Reality Labs research grant.

\clearpage %

\twocolumn[\centering \Large{\textbf{NR-NeRF --- Supplemental Material}}]

\setcounter{section}{0}
\renewcommand\thesection{S.\arabic{section}}

\setcounter{figure}{0}
\renewcommand{\thefigure}{S.\arabic{figure}}

\setcounter{table}{0}
\renewcommand{\thetable}{S.\arabic{table}}

\pdfimageresolution=20 %

\section*{Overview}

We provide further illustrations of the loss function in~Sec.~\ref{sec:loss_illustration}. 
We discuss a number of training details in~Sec.~\ref{sec:training_details}. 
Next, we provide some implementation details in~Sec.~\ref{sec:implementation_details}. 
In~Sec.~\ref{sec:data}, we describe the capture of our data. 
Sec.~\ref{sec:output_modalities}~contains details on how we visualize the additional output modalities of our method and some more results. 
We show an additional result of our ablation study in~Sec.~\ref{sec:ablation_study}. 
We provide more details on the experimental settings of our comparisons and some additional results in~Sec.~\ref{sec:comparisons}. 
Then in~Sec.~\ref{sec:extensions}, we present the extensions to multi-view data and view-dependent effects mentioned in the main paper. 
Sec.~\ref{sec:editing}~contains the scene editing tasks mentioned in the main paper. 
Sec.~\ref{sec:limitations}~contains a limitation example of our method. 
Finally,~Sec.~\ref{sec:additional_comparisons} contains some preliminary qualitative comparisons to a concurrent, non-peer-reviewed work.

\section{Loss Illustration}\label{sec:loss_illustration}

\begin{figure}[h]
\begin{center}
   \includegraphics[width=\linewidth]{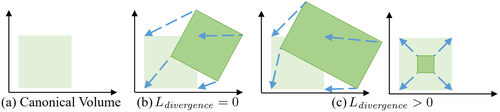}
\end{center}
   \caption{$L_\mathit{divergence}$ encourages the offsets field to (b) preserve local volume rather than (c) losing it while deforming.}
\label{fig:divergence_loss_illustration}
\end{figure}

Fig.~\ref{fig:divergence_loss_illustration} illustrates $L_\mathit{divergence}$ in 2D. 

\section{Training Details}\label{sec:training_details}

While the network weights are optimized as usual, the latent codes $\mathbf{l}_i$ are auto-decoded, i.e., they are treated as free variables that are directly optimized for, similar to network weights, instead of being regressed. 
This is based on the auto-decoding framework used in DeepSDF and earlier works~\cite{TanMayrovouniotis1995, park2019deepsdf}. 

We initialize $\{\mathbf{l}_i\}_i$ to zero vectors. 
For implementing the radiance field, we use the same architecture as in NeRF~\cite{mildenhall2020nerf}.
The ray bending network is a 5-layer MLP with 64 hidden dimensions and ReLU activations, the last layer of which is initialized with all weights set to zero.
The rigidity network is a 3-layer MLP with 32 hidden dimensions and ReLU activations, with the last layer initialized to zeros.
The output of the last layer of the rigidity network is passed though a $\operatorname{tanh}$ activation function and then shifted and rescaled to lie in $[0,1]$.
We train usually for 200k iterations with a batch of 1k randomly sampled rays. 
At training and at test time, we use 64 coarse and 64 fine samples per ray in most cases.
We use ADAM~\cite{kingma2014adam} and exponentially decay the learning rate to 10\% from the initial $5\cdot 10^{-4}$ over 250k iterations. 
For dark scenes, we found it necessary to introduce a warm-up phase that linearly increases the learning rate starting from $\frac{1}{20}$th of its original value over 1000 iterations. 
The latent codes are of the dimension 32. 
We train between six and seven hours on a single Quadro RTX 8000. 

Since we consider a variety of types of non-rigid objects/scenes and deformations, we find it necessary to use scene-specific weights for each loss term.
We have found the following ranges to be sufficient for a wide range of scenarios: $\omega_\text{rigidity}$ lies in $[0.01,0.001]$ and typically is $0.003$, $\omega_\text{offsets}$ lies in $[60,600]$ and typically is $600$, and $\omega_\text{divergence}$ lies in $[1,30]$ and typically is $3$ or $10$.
In our experience, NR-NeRF is fairly insensitive to $\omega_\text{offsets}$.
Rather rigid objects benefit from higher $\omega_\text{divergence}$, while fairly non-rigid objects need lower $\omega_\text{divergence}$.
Finally, we increase $\omega_\text{rigidity}$ whenever we find the background to be unstable. 
We start the training with each weight set to $\frac{1}{100}$th of its value, and then exponentially increase it until it reaches its full value at the end of training. 

\section{Implementation Details}\label{sec:implementation_details}
Our code is based on a faithful PyTorch~\cite{NEURIPS2019_9015} port~\cite{lin2020nerfpytorch} of the official Tensorflow NeRF code~\cite{mildenhall2020nerf}.
We use the official FFJORD implementation \cite{grathwohl2018ffjord} to estimate Eq.~5 from the main paper. %
If the camera extrinsics and intrinsics are not given, we estimate them using the Structure-from-Motion (SfM) implementation of COLMAP~\cite{schoenberger2016sfm,schoenberger2016mvs}.
We find COLMAP to be quite robust to non-rigid `outliers'.
As we are interested in estimating smooth deformations, we only apply positional encoding to the input of the canonical NeRF volume, not to the input of the ray bending network. We will make our source code available.

\section{Data}\label{sec:data}

We show results on a variety of scenes recorded with three different cameras: the Kinect Azure, a Blackmagic, and a phone camera.
Since the RGB camera of the Kinect Azure exhibits strong radial distortions along the image border, we use the manufacturer-provided intrinsics and distortion parameters to undistort the recorded RGB images beforehand.
We extract frames at 5 fps from the recordings, such that scenes usually consist of 80 to 300 images, at resolutions of $480\times 270$ (Blackmagic, and Sony XZ2) or $512\times 384$ (Kinect Azure). 

\section{Output Modalities}\label{sec:output_modalities}

\subsection{Visualizations}

\paragraph{Rigidity Scores}

In order to visualize the estimated rigidity, we need to determine the rigidity of the ray associated with a pixel.
We choose to define the rigidity of such a ray as the rigidity of the point $j$ closest to an accumulated weight $\sum_{k=0}^{j-1} \alpha_k$ of 0.5, \emph{i.e.,} closest to the median. 
In practice, this usually gives us the rigidity at the first visible surface along the ray.

\paragraph{Correspondences}
To visualize correspondences, we treat the canonical volume as an RGB cube, i.e., we treat the xyz coordinate in canonical space as an RGB color.
Since this would result in very smooth colors, we split the canonical volume into a voxel grid of $100^3$ RGB cubes beforehand.
We pick the ray point that determines the pixel color similar to the rigidity visualization.

\subsection{More Results}

\begin{figure*}

\setlength{\tabcolsep}{0.em} %
\def\arraystretch{1.} %
\resizebox{\textwidth}{!}{%
\begin{tabular}{cccccc}
  \rotatebox[origin=c]{90}{Input}
  &
  \raisebox{-0.5\height}{\includegraphics[height=.15\textheight,trim={100 0 100 0},clip]{latex/figures/additional_outputs/bag_outdoor_groundtruth_image00130.png}}
  &
  \raisebox{-0.5\height}{\includegraphics[height=.15\textheight]{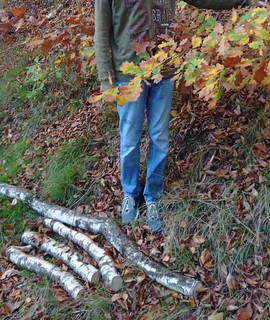}}
  &
  \raisebox{-0.5\height}{\includegraphics[height=.15\textheight]{latex/figures/additional_outputs/bush_1_groundtruth_image00069.png}}
  &
  \raisebox{-0.5\height}{\includegraphics[height=.15\textheight]{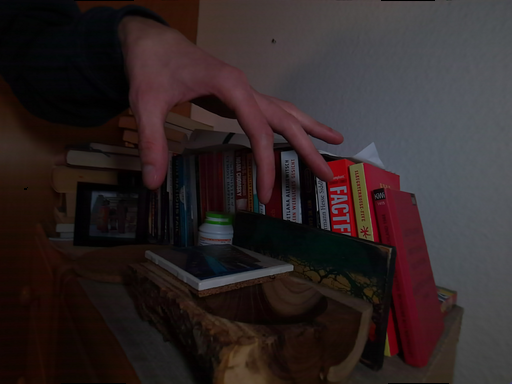}}
  &
  \raisebox{-0.5\height}{\includegraphics[height=.15\textheight]{latex/figures/additional_outputs/vlad_kicking_groundtruth_image00093.png}} 
  
  \\
  
  \rotatebox[origin=c]{90}{Color}
  &
  \raisebox{-0.5\height}{\includegraphics[height=.15\textheight,trim={100 0 100 0},clip]{latex/figures/additional_outputs/bag_outdoor_f4_ray5x64_moivd600_mdiod1_L1rig01_000129_rgb.png}}
  &
  \raisebox{-0.5\height}{\includegraphics[height=.15\textheight]{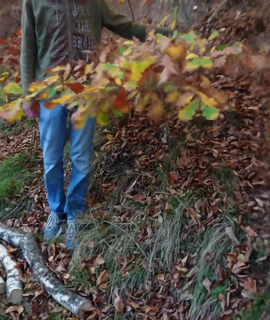}}
  &
  \raisebox{-0.5\height}{\includegraphics[height=.15\textheight]{latex/figures/additional_outputs/bush_1_f4_ray5x64_moivd600_mdiod100_L1rig6_spiral08_000068_rgb.png}}
  &
  \raisebox{-0.5\height}{\includegraphics[height=.15\textheight]{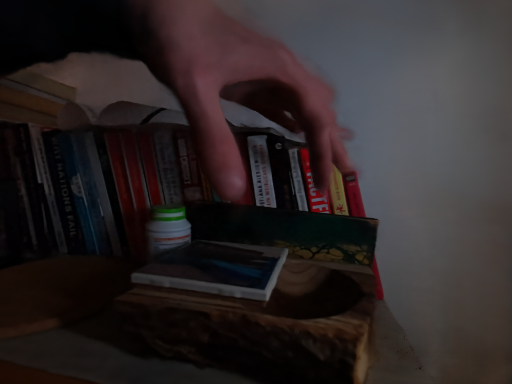}}
  &
  \raisebox{-0.5\height}{\includegraphics[height=.15\textheight]{latex/figures/additional_outputs/vlad_kicking_f4_ray5x64_moivd600_mdiod10_L1rig03_fixed10_000092_rgb.png}}
  \\
  
  \rotatebox[origin=c]{90}{Rigidity}
  &
  \raisebox{-0.5\height}{\includegraphics[height=.15\textheight,trim={100 0 100 0},clip]{latex/figures/additional_outputs/bag_outdoor_f4_ray5x64_moivd600_mdiod1_L1rig01_000129_rigidity_jet.png}}
  &
  \raisebox{-0.5\height}{\includegraphics[height=.15\textheight]{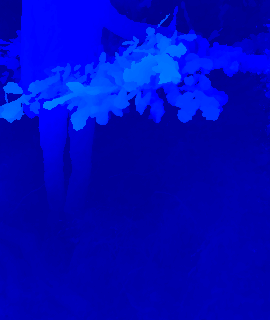}}
  &
  \raisebox{-0.5\height}{\includegraphics[height=.15\textheight]{latex/figures/additional_outputs/bush_1_f4_ray5x64_moivd600_mdiod100_L1rig6_spiral08_000068_rigidity_jet.png}}
  &
  \raisebox{-0.5\height}{\includegraphics[height=.15\textheight]{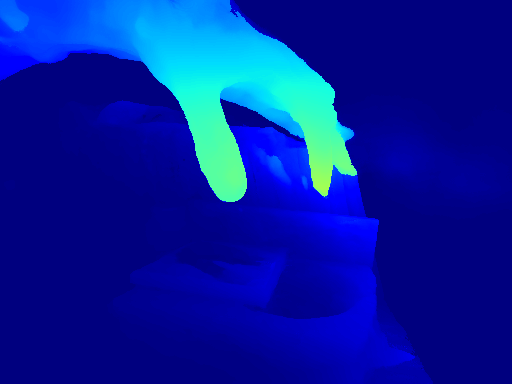}}
  &
  \raisebox{-0.5\height}{\includegraphics[height=.15\textheight]{latex/figures/additional_outputs/vlad_kicking_f4_ray5x64_moivd600_mdiod10_L1rig03_fixed10_000092_rigidity_jet.png}}
  \\

  \rotatebox[origin=c]{90}{Correspondences}
  &
  \raisebox{-0.5\height}{\includegraphics[height=.15\textheight,trim={100 0 100 0},clip]{latex/figures/additional_outputs/bag_outdoor_f4_ray5x64_moivd600_mdiod1_L1rig01_000129_correspondence_rgb.png}}
  &
  \raisebox{-0.5\height}{\includegraphics[height=.15\textheight]{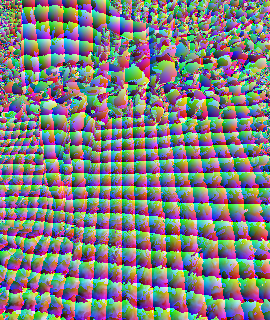}}
  &
  \raisebox{-0.5\height}{\includegraphics[height=.15\textheight]{latex/figures/additional_outputs/bush_1_f4_ray5x64_moivd600_mdiod100_L1rig6_spiral08_000068_correspondence_rgb.png}}
  &
  \raisebox{-0.5\height}{\includegraphics[height=.15\textheight]{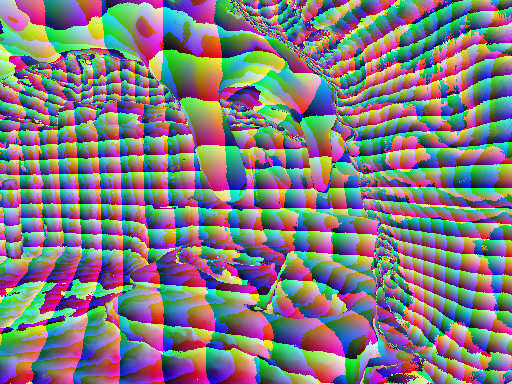}}
  &
  \raisebox{-0.5\height}{\includegraphics[height=.15\textheight]{latex/figures/additional_outputs/vlad_kicking_f4_ray5x64_moivd600_mdiod10_L1rig03_fixed10_000092_correspondence_rgb.png}}
  \\

  \rotatebox[origin=c]{90}{Canonical Volume}
  &
  \raisebox{-0.5\height}{\includegraphics[height=.15\textheight,trim={100 0 100 0},clip]{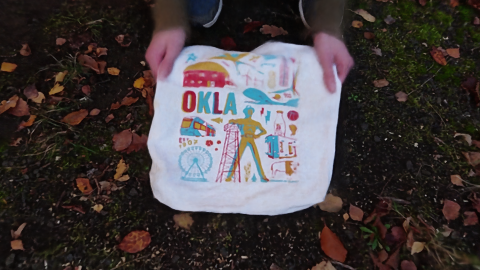}}
  &
  \raisebox{-0.5\height}{\includegraphics[height=.15\textheight]{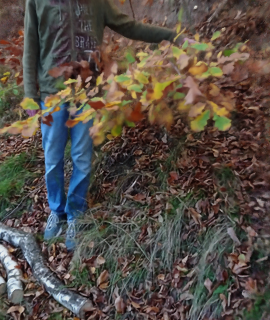}}
  &
  \raisebox{-0.5\height}{\includegraphics[height=.15\textheight]{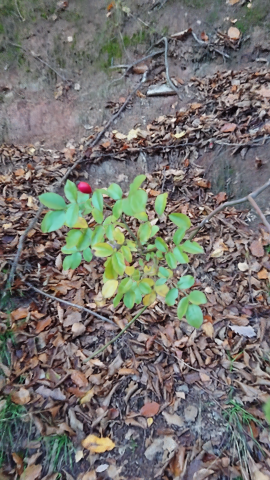}}
  &
  \raisebox{-0.5\height}{\includegraphics[height=.15\textheight]{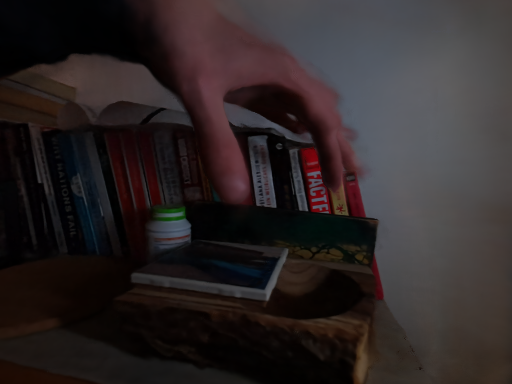}}
  &
  \raisebox{-0.5\height}{\includegraphics[height=.15\textheight]{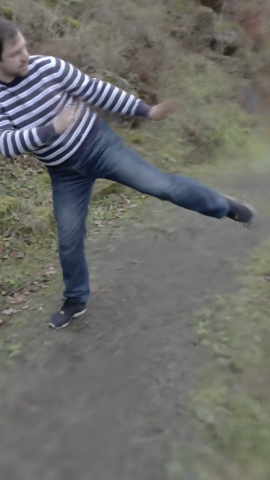}}
  \\

\textcolor{white}{.} & & & & & \\

\end{tabular}
}
  \caption{NR-NeRF can render a deformed state captured at a certain time step into a novel view. We visualize here this novel-view rendering and additional output modalities as seen from the novel view, namely rigidity scores, correspondences, and the canonical volume. The canonical volume is a plausible state of the scene and does not show baked-in deformations. We refer to the supplemental video for video results.}
  \label{fig:modalities_additional}
\end{figure*}

See Fig.~\ref{fig:modalities_additional} for more results of the output modalities of NR-NeRF.

\paragraph{Canonical Volume}
Since the canonical volume is not supervised directly, it is conceivable that it could have baked-in deformations.
The last row of Fig.~\ref{fig:modalities_additional} contains renderings of the canonical volume without any ray bending applied.
The canonical volume is a plausible state of the scene and does not show baked-in deformations.
We thus find it to be sufficient to bias the optimization towards a desirable canonical volume by initializing the ray bending network to an identity map and by our regularization losses.

\section{Ablation Study}\label{sec:ablation_study}

\begin{figure}
\setlength{\tabcolsep}{0.3em} %
\def\arraystretch{0.3} %
\begin{tabular}{ccc}
  
  \includegraphics[width=.15\textwidth,trim={120 0 50 0},clip]{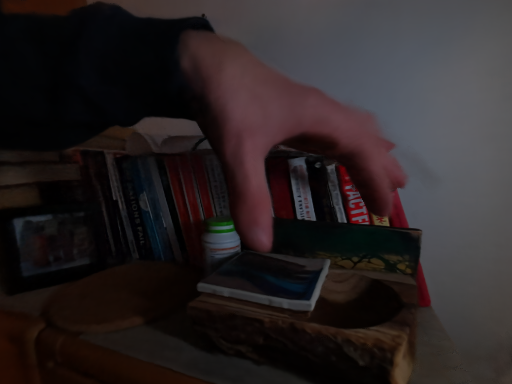}
  &
  \includegraphics[width=.15\textwidth,trim={120 0 50 0},clip]{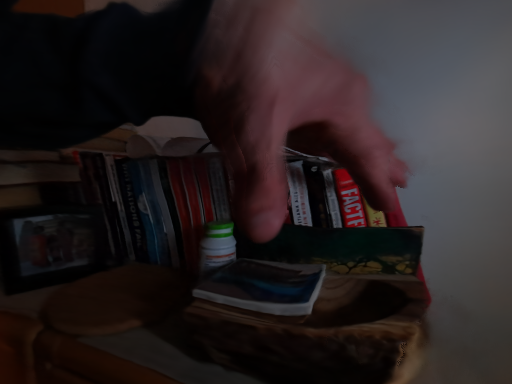}
  &
  \includegraphics[width=.15\textwidth,trim={120 0 50 0},clip]{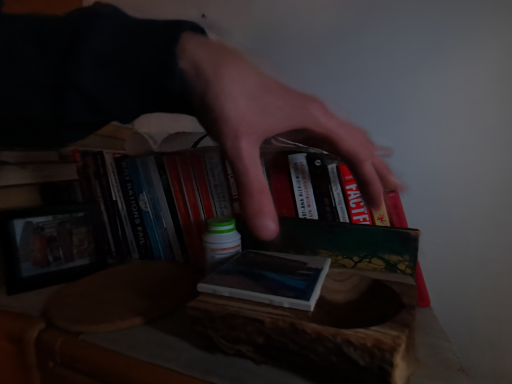}
  \\
  
  \small Without $L_\mathit{divergence}$ &  \small No regularization &  \small Ours
  
\end{tabular}
  \caption{Ablation Study. We render the input scene into a novel view to determine the stability of the non-rigid objects. We show the results of removing the divergence loss, all three regularization losses, and none of the losses. 
  }
  \label{fig:divergence_qualitative_additional}
\end{figure}

Fig.~\ref{fig:divergence_qualitative_additional} contains an additional ablation study result that demonstrates the importance of the divergence regularization for foreground stability.

\section{Comparisons}\label{sec:comparisons}

In this section, we provide more details on the experimental settings of the comparisons.

\subsection{Prior Work and Baseline}

We start with the trivial baseline of rigid NeRF~\cite{mildenhall2020nerf}, which cannot handle dynamic scenes.
We consider two variants: view-dependent rigid NeRF, as in the original method~\cite{mildenhall2020nerf}, and view-independent rigid NeRF, where we remove the view-direction conditioning.

We next introduce \emph{na\"ive NR-NeRF}, which adds na\"ive support for dynamic scenes to rigid NeRF: We condition the neural radiance fields volume on the latent code.
Thus, for latent code $\mathbf{l}_i$, we have $(\mathbf{c}, o)=\mathbf{v}(\mathbf{x}, \mathbf{l}_i)$.
This allows the neural radiance fields volume to output time-varying color and occupancy.
Unlike NR-NeRF's ray bending, na\"ive NR-NeRF does not have an explicit, separate deformation model.
Instead, the volume needs to account for appearance, geometry and deformation at once.
Note that for test images $i$, we do backpropagate gradients into the corresponding latent code $\textbf{l}_i$.

Finally, we compare to Neural Volumes~\cite{lombardi2019neural}, for which we use the official code release.
We use the standard settings as a starting point, but set the number of training iterations to $100k$, which leads to a training time of about two days on an RTX8000 GPU.
Neural Volumes uses an image encoder to regress a latent code that conditions the geometry, appearance and deformation regression on the current time step.
Since this design assumes a multi-view setup, we need to adapt it to our monocular setting.
Instead of picking three fixed camera views that are always input into the encoder, we input the single image of the current time step.
In particular, at test time, we input the test image.
Since we do not have access to a background image, we set the estimated background image to an all-black image.
We furthermore consider two variations: (1) following the original Neural Volumes method, the geometry and appearance template is conditioned on the latent code (\textit{NV}), and (2) the geometry and appearance template is independent of the latent code (\textit{modified NV}).
In the latter case, the latent code only conditions the warp field, which is similar to our method.

\subsection{Training/Test Split}
For quantitative evaluation, we require a test set.
In the comparison section, we split the images into training and test images by partitioning the temporally-ordered images into consecutive blocks, each of length 16.
The first twelve images of each block are used as training data, while the remaining four are used for testing.
In our setting, test images still require corresponding latent codes to represent the deformations.
Therefore, we treat test images like training images except that we do not backpropagate into the canonical volume or the ray bending network.
However, we do use gradients from test images to optimize the corresponding latent codes. 
(Note that test images \emph{solely} influence test latent codes, as is typical for auto-decoding~\cite{park2019deepsdf}.
\footnote{The only tweak we add is that we align the optimization landscapes of the training and test latent codes by optimizing the test latent codes jointly with the training latent codes during training. 
This does not lead to any information leakage from the test images to any component except for the test latent codes.})
All other results shown in this paper, outside of the comparisons, treat all images as training images since we train scene-specific networks.
Furthermore, qualitative baseline results in the supplemental video show both training and test time steps.

\subsection{Additional Results}

\begin{figure*}

\setlength{\tabcolsep}{0.05em} %
\def\arraystretch{0.1} %

\begin{tabular}{cccccccc}

  &
  Input
  &
  Ours
  &
  Na\"ive NR-NeRF
  &
  Rigid NeRF
  &
  Rigid NeRF
  &
  Neural Volumes
  &
  Neural Volumes
  \\
  
  &
  &
  &
  &
  (view-dep.)
  &
  (not view-dep.)
  &

  &
  (modified)
  \\
  
  &
  \includegraphics[width=.14\textwidth]{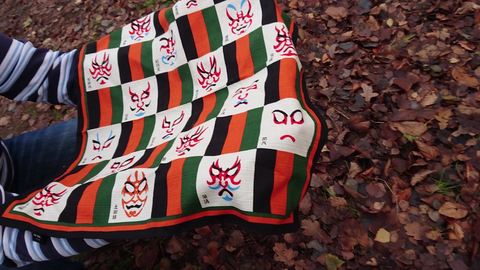}
  &
  \includegraphics[width=.14\textwidth]{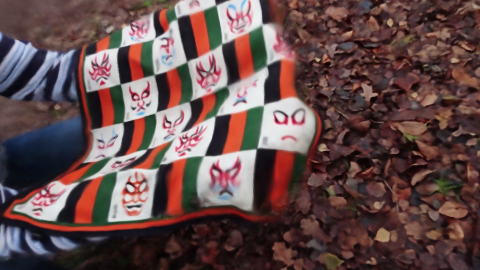}
  &
  \includegraphics[width=.14\textwidth]{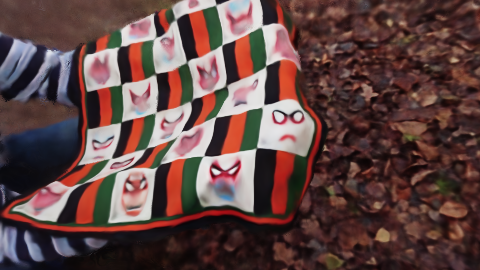}
  &
  \includegraphics[width=.14\textwidth]{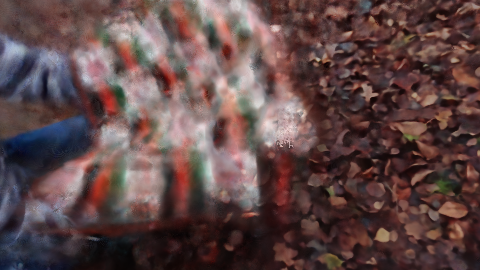}
  &
  \includegraphics[width=.14\textwidth]{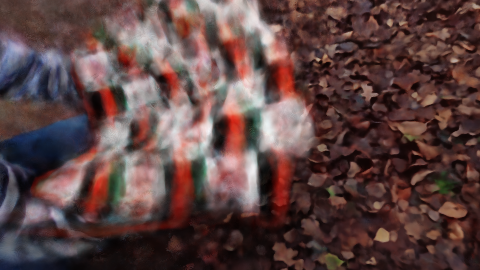}
  &
  \includegraphics[width=.14\textwidth]{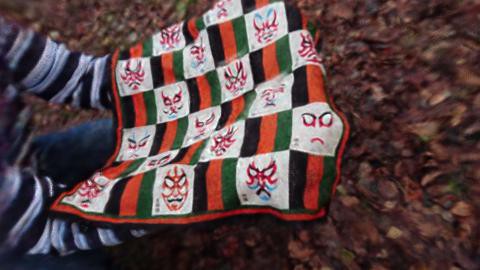}
  &
  \includegraphics[width=.14\textwidth]{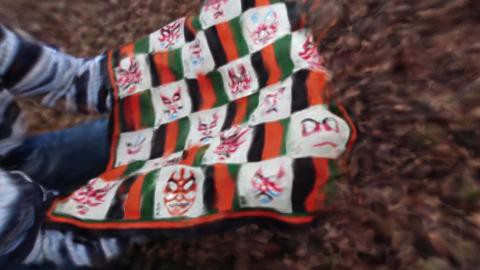}
  \\
  
  &
  
  &
  \includegraphics[width=.14\textwidth]{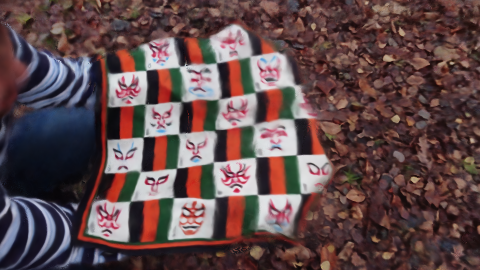}
  &
  \includegraphics[width=.14\textwidth]{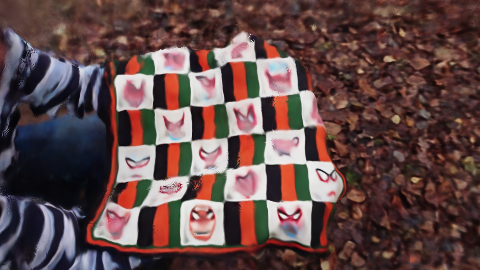}
  &
  \includegraphics[width=.14\textwidth]{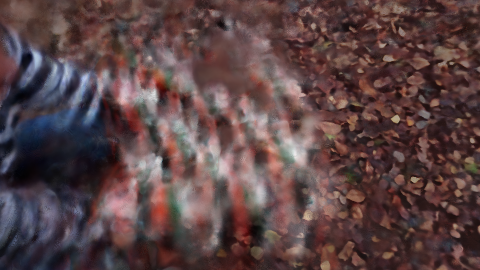}
  &
  \includegraphics[width=.14\textwidth]{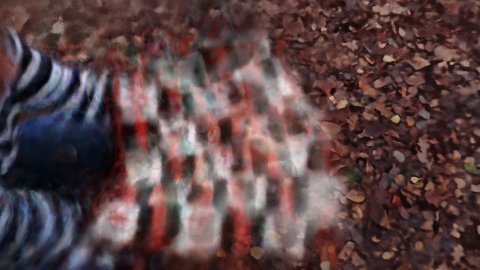}
  &
  \includegraphics[width=.14\textwidth]{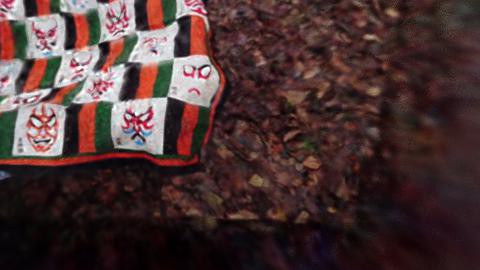}
  &
  \includegraphics[width=.14\textwidth]{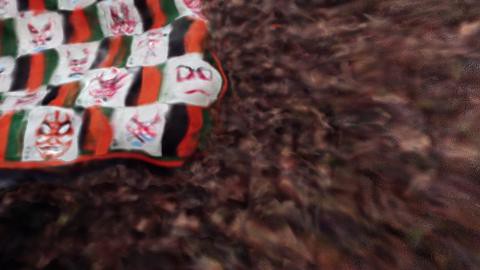}
  \\

  &
  \includegraphics[width=.14\textwidth]{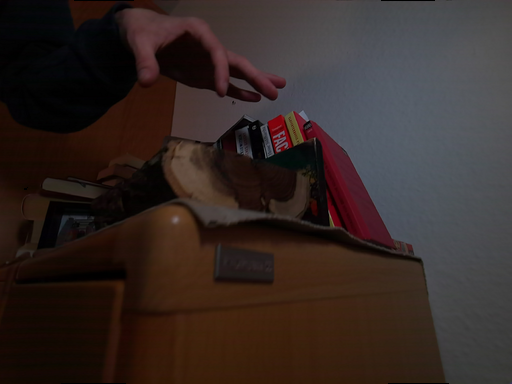}
  &
  \includegraphics[width=.14\textwidth]{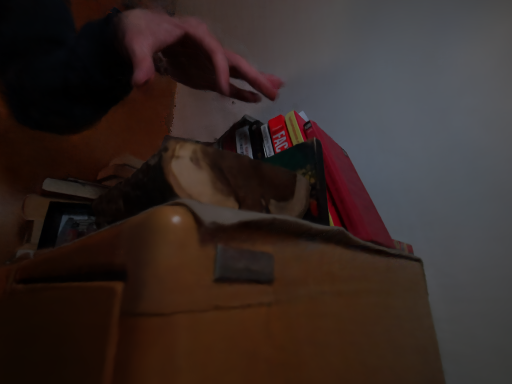}
  &
  \includegraphics[width=.14\textwidth]{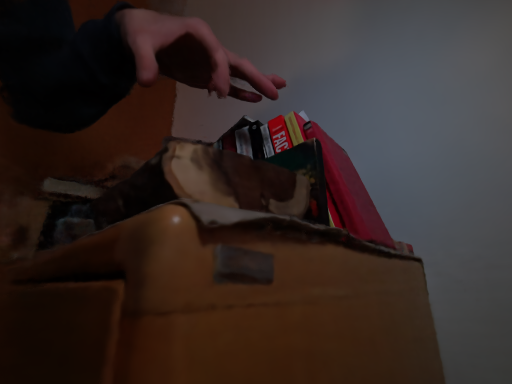}
  &
  \includegraphics[width=.14\textwidth]{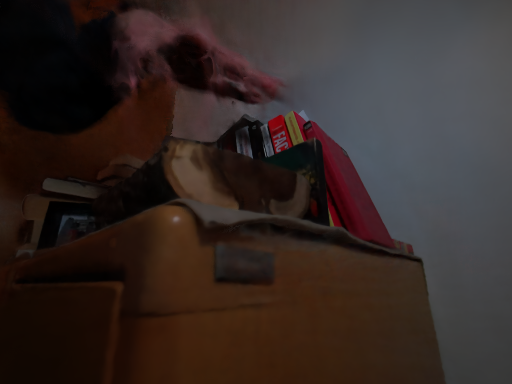}
  &
  \includegraphics[width=.14\textwidth]{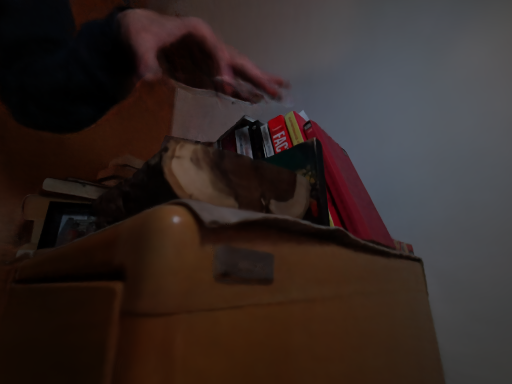}
  &
  \includegraphics[width=.14\textwidth]{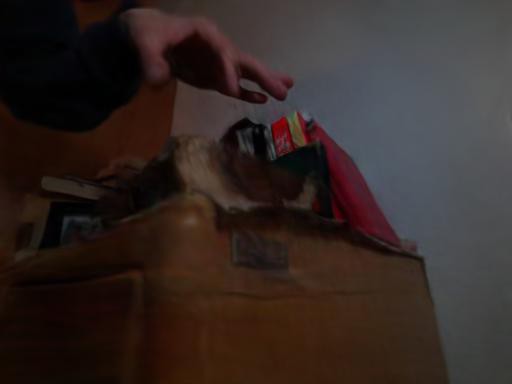}
  &
  \includegraphics[width=.14\textwidth]{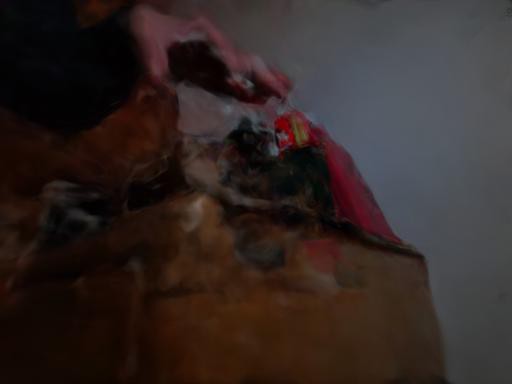}
  \\
  
  &
  
  &
  \includegraphics[width=.14\textwidth]{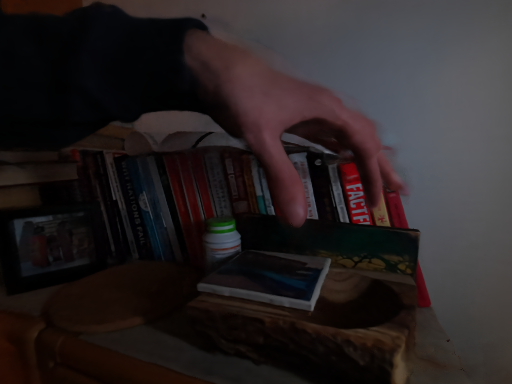}
  &
  \includegraphics[width=.14\textwidth]{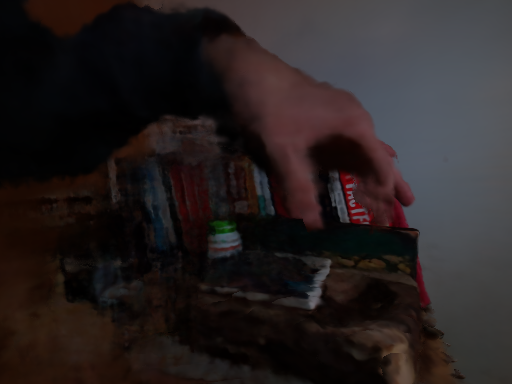}
  &
  \includegraphics[width=.14\textwidth]{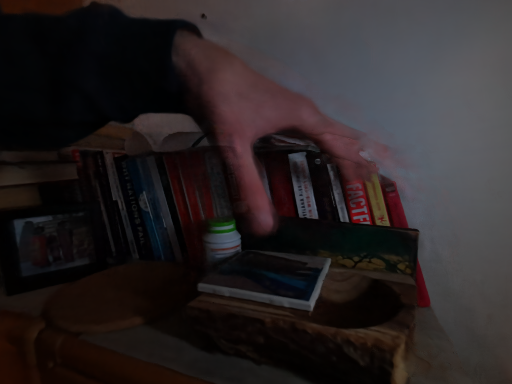}
  &
  \includegraphics[width=.14\textwidth]{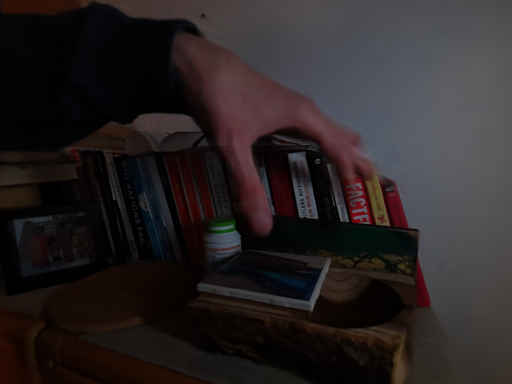}
  &
  \includegraphics[width=.14\textwidth]{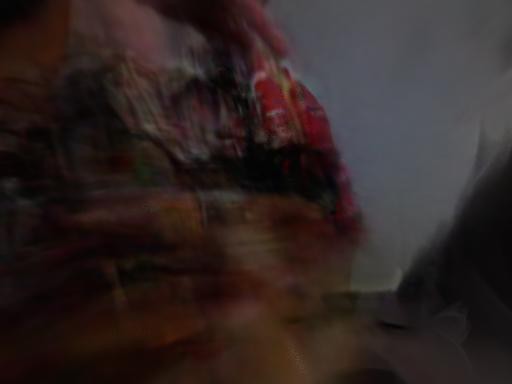}
  &
  \includegraphics[width=.14\textwidth]{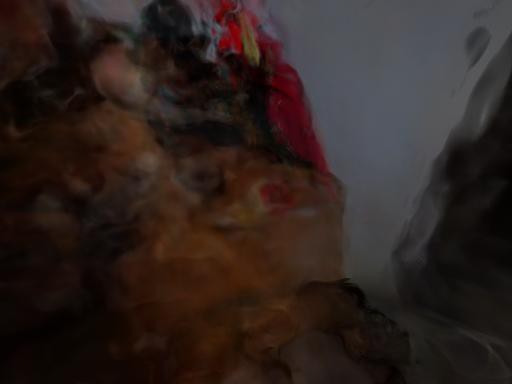}
  \\

  &
  \includegraphics[width=.14\textwidth]{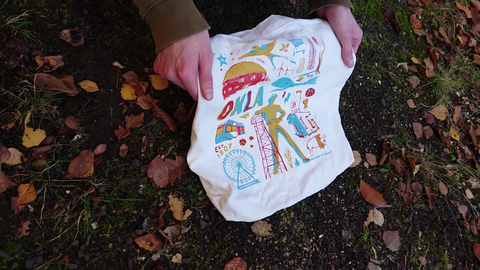}
  &
  \includegraphics[width=.14\textwidth]{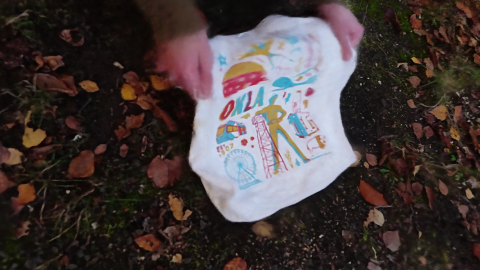}
  &
  \includegraphics[width=.14\textwidth]{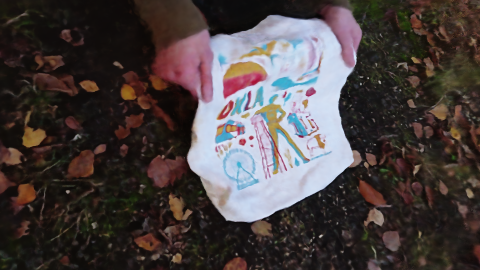}
  &
  \includegraphics[width=.14\textwidth]{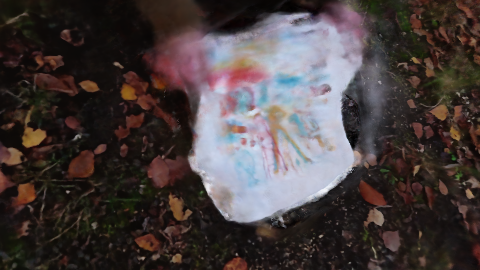}
  &
  \includegraphics[width=.14\textwidth]{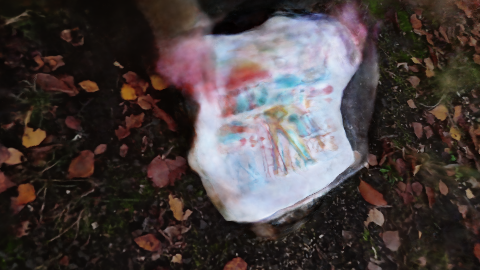}
  &
  \includegraphics[width=.14\textwidth]{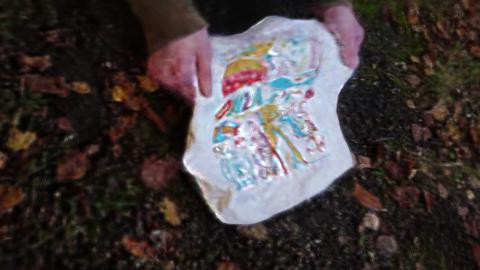}
  &
  \includegraphics[width=.14\textwidth]{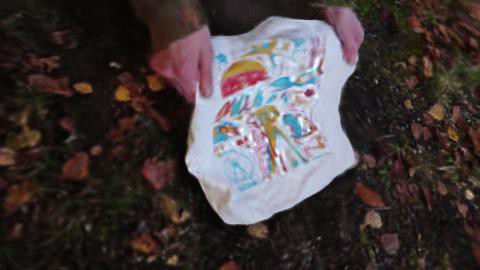}
  \\
  
  &
  
  &
  \includegraphics[width=.14\textwidth]{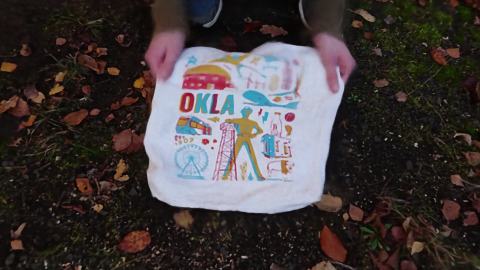}
  &
  \includegraphics[width=.14\textwidth]{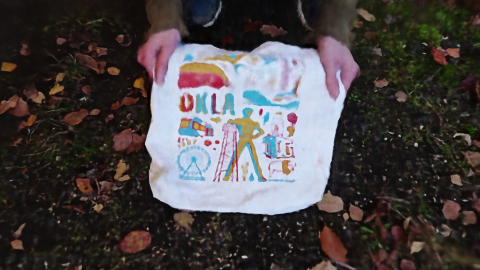}
  &
  \includegraphics[width=.14\textwidth]{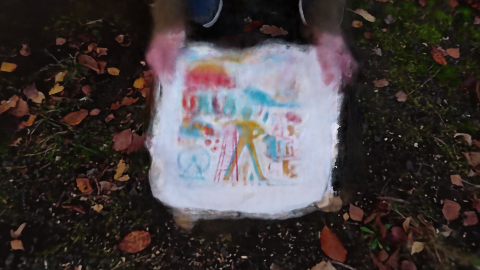}
  &
  \includegraphics[width=.14\textwidth]{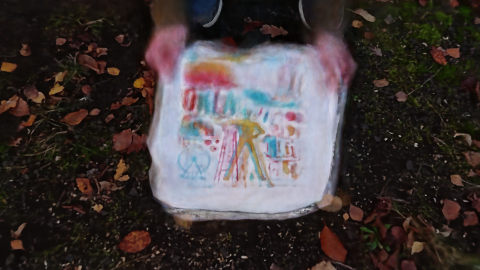}
  &
  \includegraphics[width=.14\textwidth]{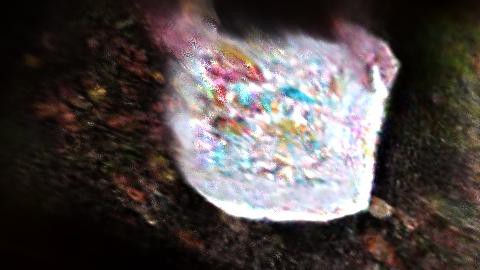}
  &
  \includegraphics[width=.14\textwidth]{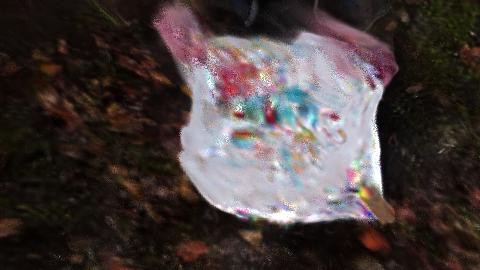}

\end{tabular}
  \caption{We show one time step each from each sequence and compare input reconstruction quality (first row) and novel view synthesis quality (second row).
  } 
  \label{fig:qualitative_comparison_additional}
\end{figure*}

\begin{figure}

\setlength{\tabcolsep}{0.em} %
\def\arraystretch{0.7} %
\begin{tabular}{cccc}

\multicolumn{1}{l}{\large{a)}} & & & 
\\

  \multicolumn{2}{c}{\includegraphics[width=.45\columnwidth]{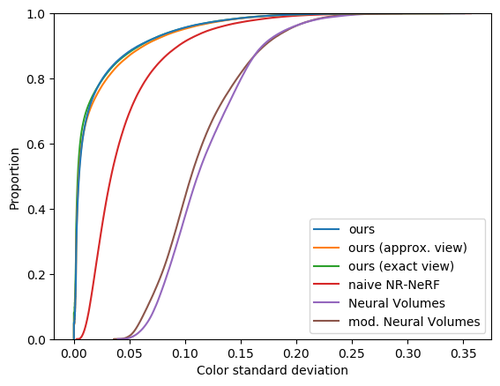}}
  &
  \multicolumn{2}{c}{\includegraphics[width=.45\columnwidth]{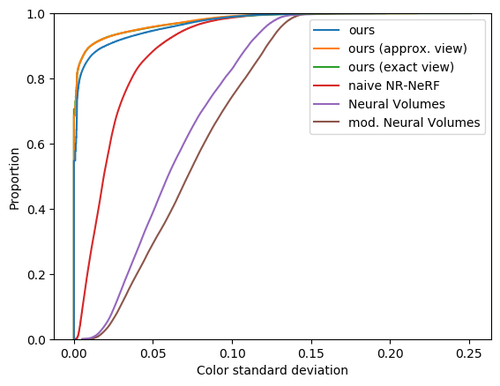}}
  \\
 
  \multicolumn{2}{c}{Left Scene} & \multicolumn{2}{c}{Right Scene}\\
 
 \multicolumn{1}{l}{\large{b)}} & & & 
\\
  
  \multicolumn{4}{c}{\includegraphics[width=.45\columnwidth]{latex/figures/ablation/jet_color_spectrum_annotated2.png}} \\

  \includegraphics[width=.28\columnwidth]{latex/figures/stick_still_2_f4_ray5x64_moivd600_mdiod30_L1rig2_split12_4_standard_deviations.png}
  &
  \includegraphics[width=.21\columnwidth]{latex/figures/hand_thumb_index_easier_f4_ray5x64_moivd60_mdiod3_L1rig003_split12_4_standard_deviations.png}~~
  &
  \includegraphics[width=.28\columnwidth]{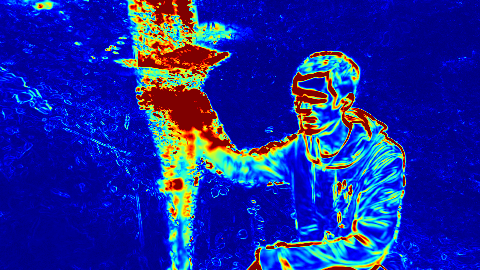}
  &
  \includegraphics[width=.21\columnwidth]{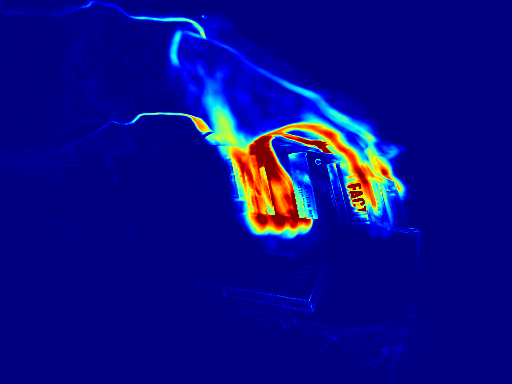}
  \\  
  \multicolumn{2}{c}{\small Ours}
  &
    \multicolumn{2}{c}{\small Ours (approx. view)}
  \\
  \tiny\color{white} . & & & \\

  \includegraphics[width=.28\columnwidth]{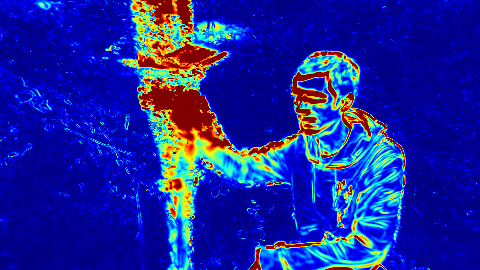}
  &
  \includegraphics[width=.21\columnwidth]{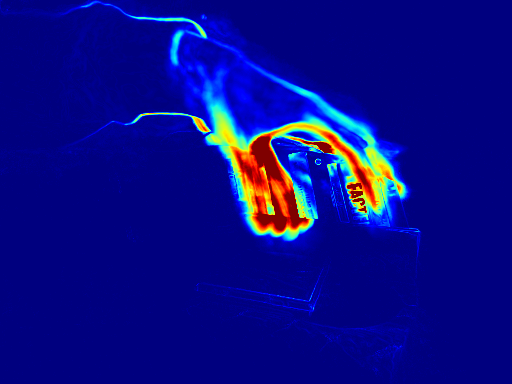}~~
  &
  \includegraphics[width=.28\columnwidth]{latex/figures/stick_still_2_f4_timecondbaseline_split12_4_standard_deviations.png}
  &
  \includegraphics[width=.21\columnwidth]{latex/figures/hand_thumb_index_easier_f4_timecondbaseline_split12_4_standard_deviations.png}
  \\
  \multicolumn{2}{c}{\small Ours (exact view)}
  &
    \multicolumn{2}{c}{\small Na\"ive NR-NeRF}
  \\
  \tiny\color{white} . & & & \\

  \includegraphics[width=.28\columnwidth]{latex/figures/stick_still_2_f4_neural_volumes_nonshared_standard_deviations.png}
  &
  \includegraphics[width=.21\columnwidth]{latex/figures/hand_thumb_index_easier_f4_neural_volumes_nonshared_standard_deviations.png}~~
  &
    \includegraphics[width=.28\columnwidth]{latex/figures/stick_still_2_f4_neural_volumes_shared_standard_deviations.png}
  &
  \includegraphics[width=.21\columnwidth]{latex/figures/hand_thumb_index_easier_f4_neural_volumes_shared_standard_deviations.png}
  \\
  \multicolumn{2}{c}{\small Neural Volumes }
  &
    \multicolumn{2}{c}{\small Neural Volumes (modified)}
  \\

\end{tabular}%

  \caption{Background Stability. We quantify the difference in background stability between our method, its variants with view dependence, na\"ive NR-NeRF, and Neural Volumes. To that end, we render all test time steps of the input sequence into a fixed novel view and compute the standard deviation of each pixel's color across time to measure color changes and hence background stability. \textbf{a)}~We show cumulative plots across all pixels, where NR-NeRF and its variants (left-most curves) have the most stable background. \textbf{b)}~We then show how those instabilities are distributed in the scene. The results of NR-NeRF and its variants show the least instability in the background.
  }
  \label{fig:comparison_background_instability_additional}
\end{figure}

See Fig.~\ref{fig:qualitative_comparison_additional} for more qualitative results and Fig.~\ref{fig:comparison_background_instability_additional} for quantitative results on background stability under novel views. 

\section{Extensions}\label{sec:extensions}

As mentioned in Sec.~4 from the main paper, we can extend our approach easily to work with multi-view data and view-dependent effects.

\subsection{Multi-View Data}
Our approach naturally handles multi-view data. 
Although we mainly work with monocular data, we can use multi-view data to investigate the upper quality bound of our approach under ideal real-world conditions.
\paragraph{Method}
Instead of each image having its own time step and hence latent code, images taken at the same time step share the same latent code.
This ensures that the canonical volume deforms consistently within each time step.

\paragraph{Data and Settings}
We use a multi-view dataset that has 16 camera pairs evenly distributed around the scene, which sufficiently constrains the optimization such that we find the training to not need any regularization losses.
We train at the original resolution of $5120\times 3840$ for 2 million training iterations with 4096 rays per batch and 256 coarse and 128 fine samples.
These highest-quality settings lead to a training time of 11 days on 4 RTX8000 GPUs, and a rendering time of about 10 minutes per frame on the same hardware.

\paragraph{Results}
See Fig.~\ref{fig:multi_view_view_dependence} and the supplemental video for results on five consecutive time steps.

\subsection{View Dependence}
We can optionally add view-dependent effects, like specularities, into our model.
\paragraph{Method}
Determining the view direction or ray direction is not as trivial as for the straight rays.
Instead, we need to calculate the direction in which the bent ray passes through a point in the canonical volume.
We consider two options of doing so: exact and slower, or approximate and faster.

\emph{Exact:} We obtain the direction of the bent ray $\tilde{\mathbf{r}}$ at a point $\tilde{\mathbf{r}}(j)$ via the chain rule as $\nabla_j\tilde{\mathbf{r}}(j) = \frac{\partial \tilde{\mathbf{r}}(j)}{\partial \bar{\mathbf{r}}(j)} \cdot \frac{\partial \bar{\mathbf{r}}(j)}{\partial j} = J \cdot \mathbf{d}$, where $J$ is the $3\times 3$ Jacobian and $\mathbf{d}$ is the direction of the straight ray.
We compute $J$ via three backward passes (one for each output dimension), which is computationally expensive.

\emph{Approximate:} To reduce computation, we can approximate the direction at the ray sample via finite differences as the normalized difference vector between the current point $\tilde{\mathbf{r}}(j)$ and the previous point $\tilde{\mathbf{r}}(j-1)$ along the bent ray (which is closer to the camera).

\begin{figure*}

\begin{tabular}{cccc}
  \multicolumn{2}{c}{
  \includegraphics[width=.425\textwidth]{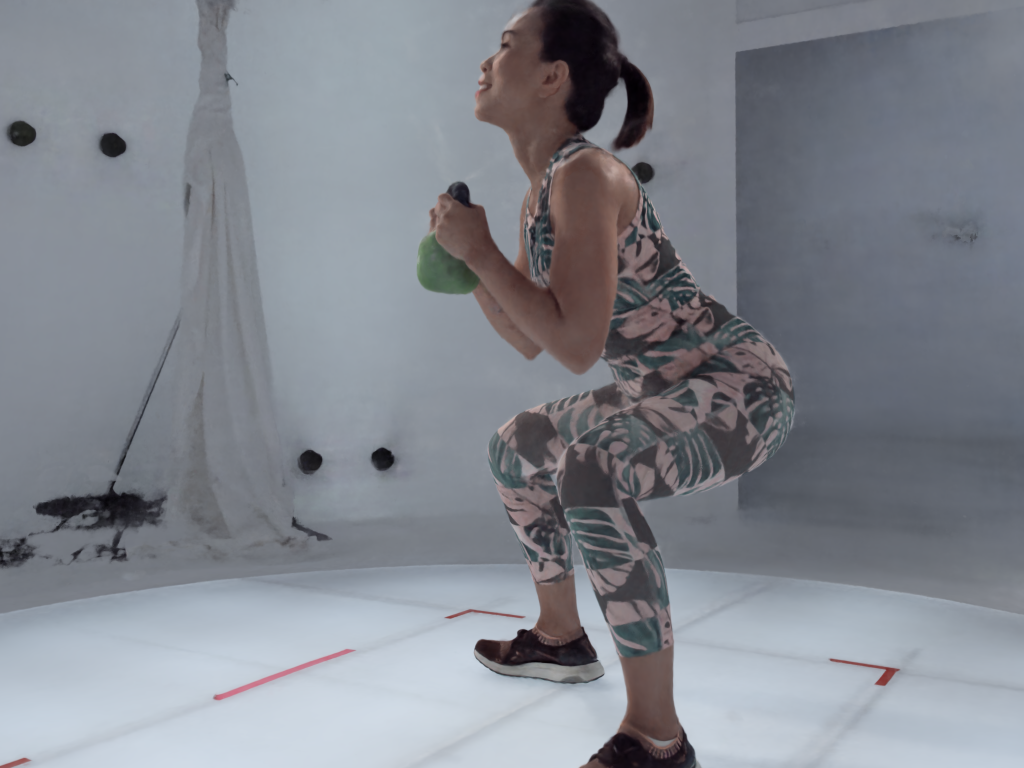}}
  &
  \multicolumn{2}{c}{
  \includegraphics[width=.425\textwidth]{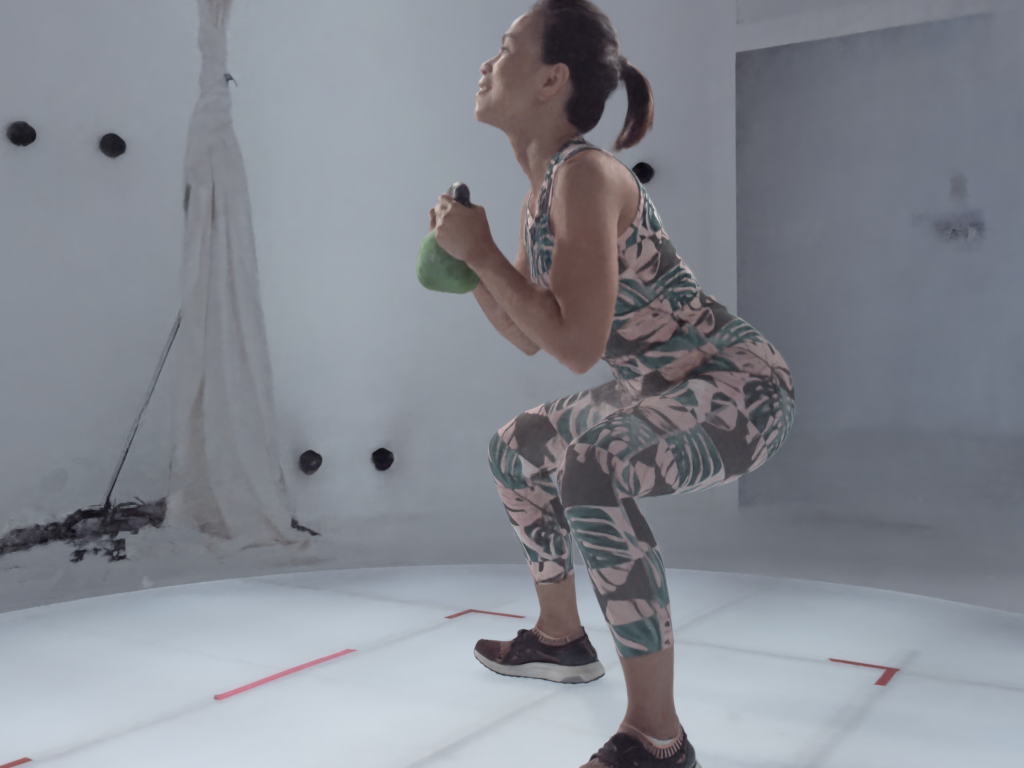}}
  \\
  \multicolumn{2}{c}{
  \includegraphics[width=.425\textwidth]{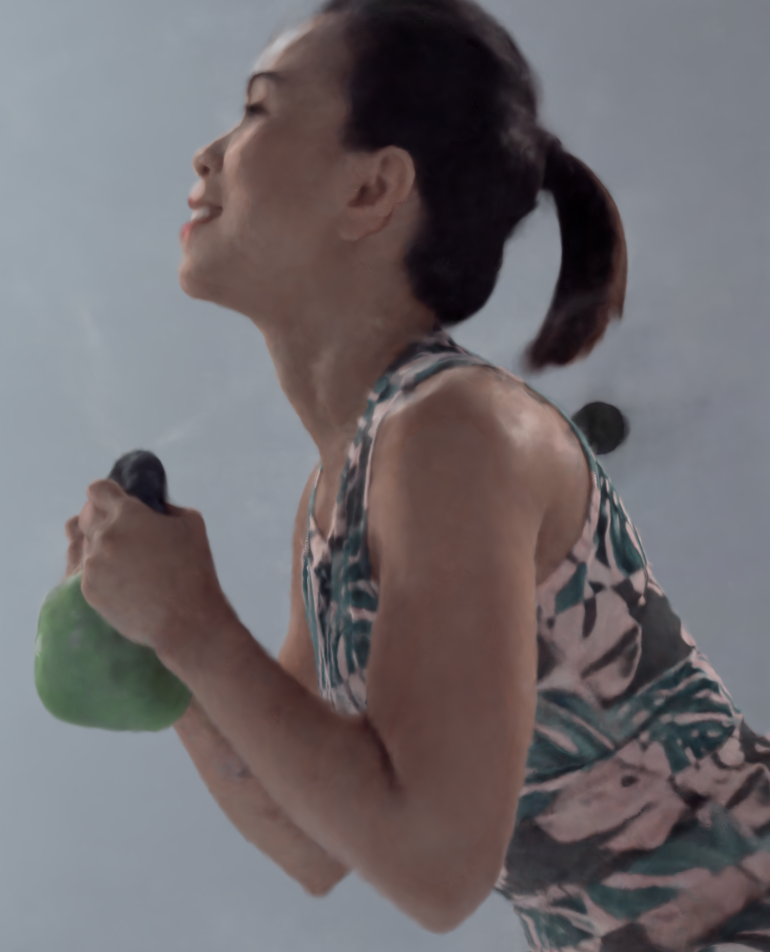}}
  &
  \multicolumn{2}{c}{
  \includegraphics[width=.425\textwidth]{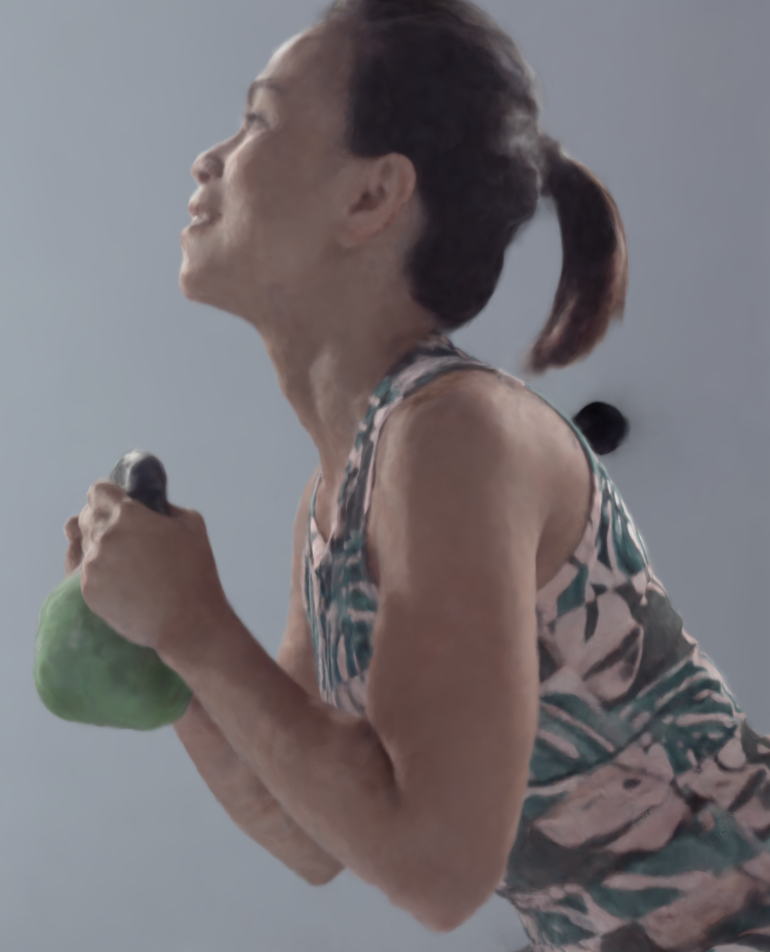}}
  \\
  \includegraphics[height=.25\textheight]{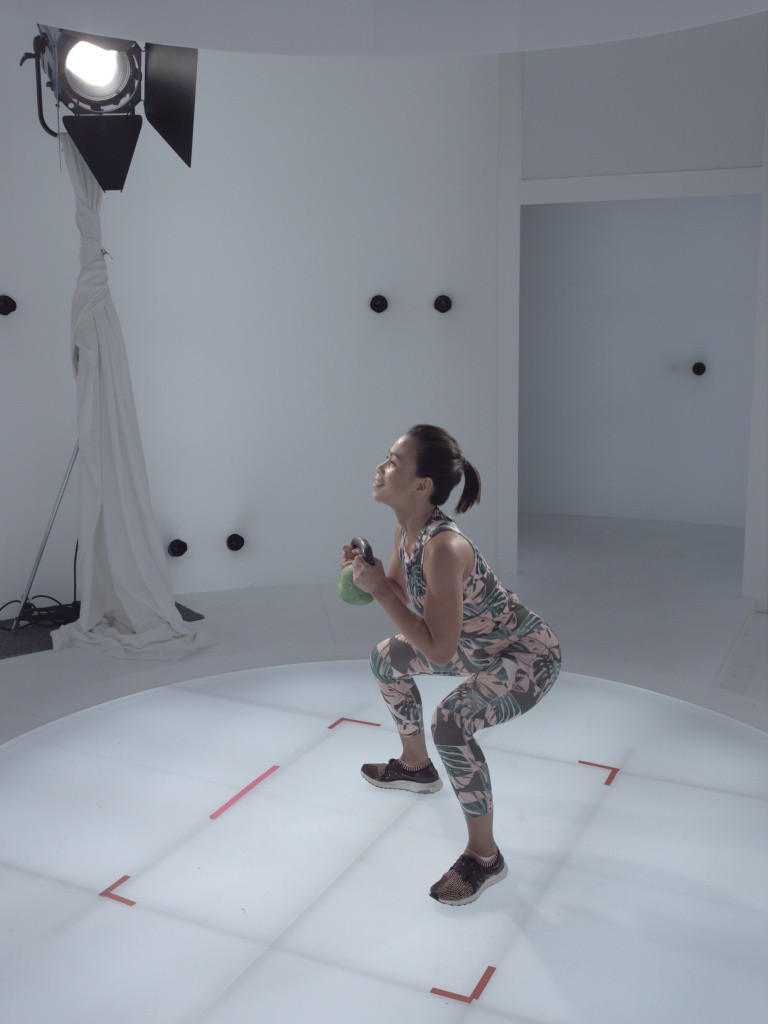}
  &
  \includegraphics[height=.25\textheight]{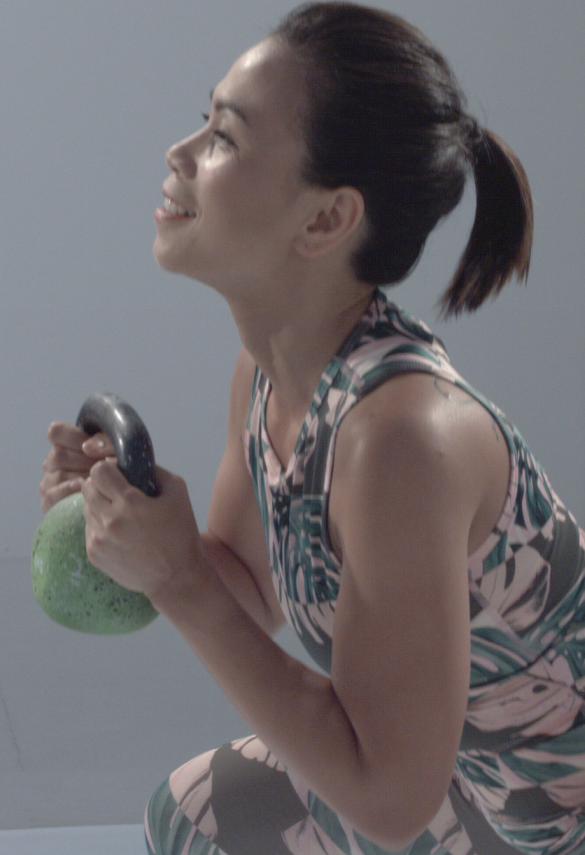}
  &
  \includegraphics[height=.25\textheight]{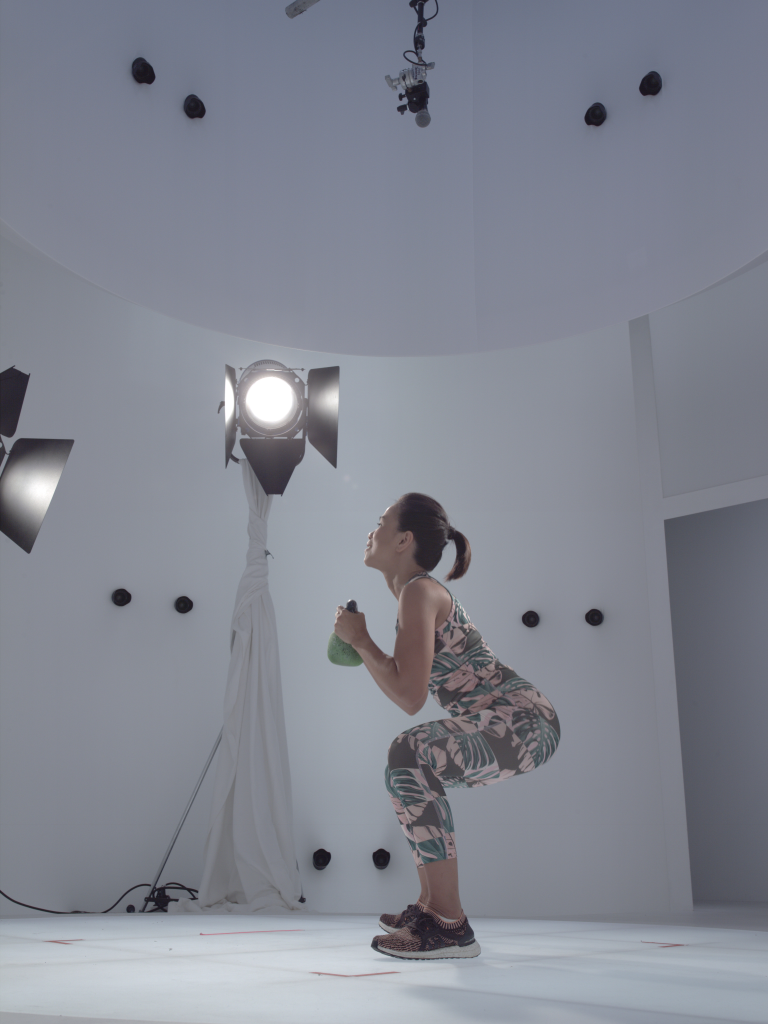}
  &
  \includegraphics[height=.25\textheight]{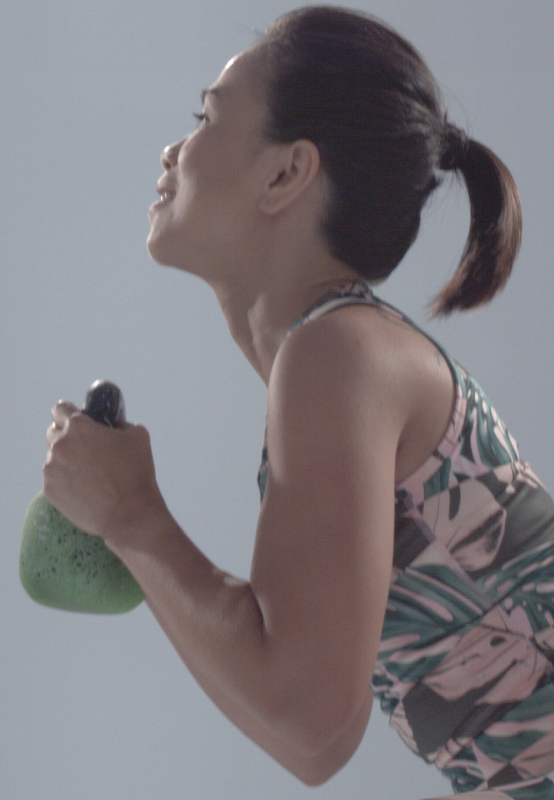}

\end{tabular}
  \caption{We explore the upper quality bound of our proposed method using a highly controlled multi-view setting. We can extend our method such that it handles view-dependent effects. Results on the left are without view dependence, while those on the right are with view dependence. We show a full rendering by NR-NeRF (first row), zoom-ins thereof (second row), and input images from the two closest input cameras.}
  \label{fig:multi_view_view_dependence}
\end{figure*}

\paragraph{Results}
On multi-view data, conditioning on the viewing direction reduces the presence of subtle, smoke-like artifacts, which the canonical volume typically employs to model view-dependent effects without view conditioning. 
This is especially visible for the specularities on the face and the handle of the kettlebell. 
Without view-dependent effects, the reconstructed face still appears to exhibit specularities, but these are \emph{incorrectly} modeled via smoke-like artifacts in the surrounding air.
See Fig.~\ref{fig:multi_view_view_dependence} and the supplemental video for results.

\begin{figure}
\centering
\setlength{\tabcolsep}{0.1em} %
\def\arraystretch{0.3} %
\begin{tabular}{ccc}
  \includegraphics[width=.15\textwidth]{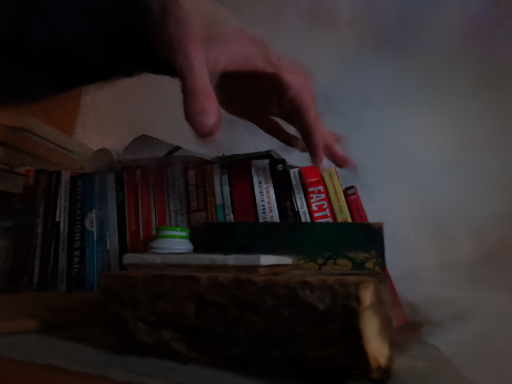} 
  &
  \includegraphics[width=.15\textwidth]{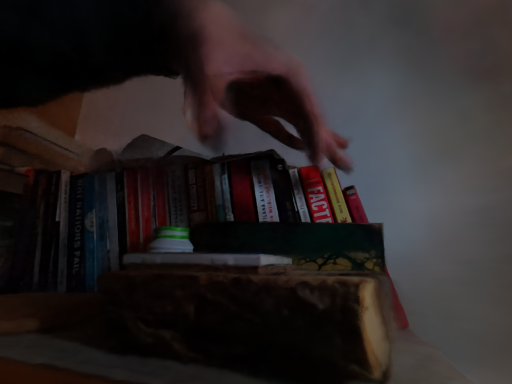} 
  &
  \includegraphics[width=.15\textwidth]{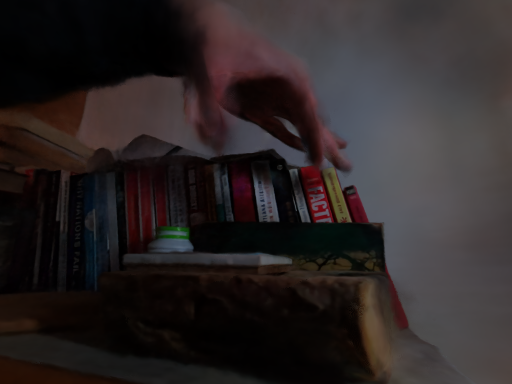}
  \\
  \small no (default) & \small approximate & \small exact
\end{tabular}
  \caption{While NR-NeRF extended with view-dependent effects (approximate or exact) gives similar results to the default NR-NeRF for many monocular scenes, we sometimes observe artifacts for difficult novel views. }
  \label{fig:monocular_view_dependence}
\end{figure}

For quantitative results on monocular sequences, see Tab.~\ref{tab:quantitative_additional}. 
However, as Fig.~\ref{fig:monocular_view_dependence} shows, we find our formulation to lead to artifacts in some cases.
We hypothesize that the combination of both significant motion and novel views significantly different from input views is too underconstrained for view-dependent effects.
For example, %
non-rigid NeRF might incorrectly overfit to subtle correlations between deformation and camera position at training time.
However, we want to emphasize that better formulations and regularization in future work may make view-dependent effects work in these challenging scenarios.

\begin{table}[]
\resizebox{\columnwidth}{!}{%
\begin{tabular}{l|c|c|c|c|c|c|c|c|}
&  Ours &  Ours &  Ours &  Na\"ive  &  Rigid &  Rigid & NV & NV\\
& \small  & \small (appx.) & \small (exact) &  & \small (cond.) & \small (no cond.) & & \small (mod.)\\
\cline{1-9} 
PSNR & 24.70 & $\mathit{25.15}$ & 25.07 & $\mathbf{25.83}$ & 22.24 & 21.88 & 14.13 & 14.10\\
SSIM & 0.758 & $\mathbf{0.766}$ & $\mathit{0.765}$ & 0.738 & 0.662 & 0.659 & 0.259 & 0.263 \\
LPIPS & 0.197 & $\mathit{0.191}$ & $\mathbf{0.190}$ & 0.226 & 0.309 & 0.313 & 0.580 & 0.583 \\
\hline
\end{tabular}%
}
\caption{Quantitative Results Averaged Across Scenes. We evaluate our method (1) without view conditioning, (2) with approximate view conditioning, and (3) with exact view conditioning, na\"ive NR-NeRF, rigid NeRF~\cite{mildenhall2020nerf} (1) with view conditioning and (2) without view conditioning, and Neural Volumes~\cite{lombardi2019neural} (1) without and (2) with modifications. For PSNR and SSIM \cite{ssim2004}, higher is better. For LPIPS \cite{zhang2018perceptual}, lower is better. As in Table 1 in the main document, we use 18 scenes here, with an average length of 146 frames and a minimum of 41 and a maximum of 453 frames.}
\end{table}\label{tab:quantitative_additional}

\section{Simple Scene Editing}\label{sec:editing}

We can manipulate the learned model in further simple ways: foreground removal, temporal super-sampling, deformation exaggeration or dampening, and forced background stabilization. 
Having discussed only foreground removal in the main paper due to space constraints, we here present the other editing tasks.

\subsection{Time Interpolation}
We can linearly interpolate between consecutive time steps to enable temporal super-resolution since NR-NeRF optimizes a latent code $\mathbf{l}_i$ for every time step $i$.
We refer to the supplemental video for results.

\begin{figure}
\setlength{\tabcolsep}{0.em} %
\def\arraystretch{0.5} %
\begin{tabular}{cccccc}
  \includegraphics[width=.165\columnwidth]{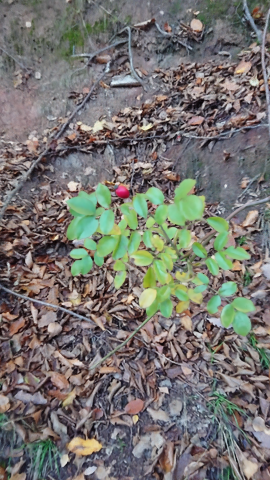}
  &
  \includegraphics[width=.165\columnwidth]{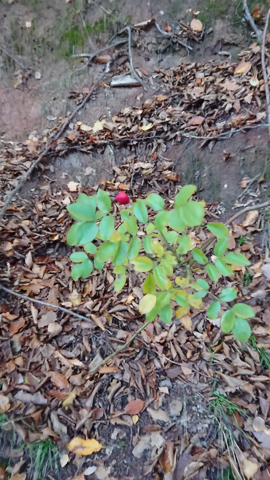}
  &
  \includegraphics[width=.165\columnwidth]{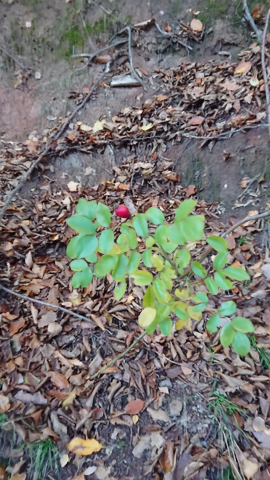}
  &
  \includegraphics[width=.165\columnwidth]{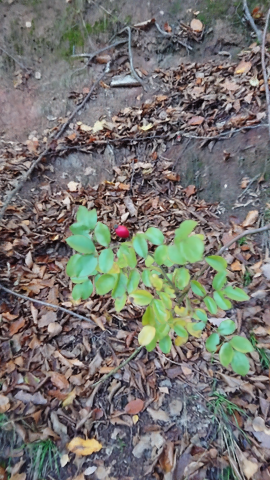}
  &
  \includegraphics[width=.165\columnwidth]{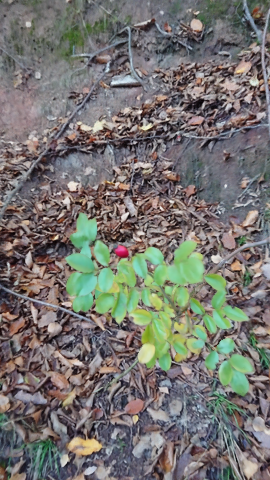}
    &
  \includegraphics[width=.165\columnwidth]{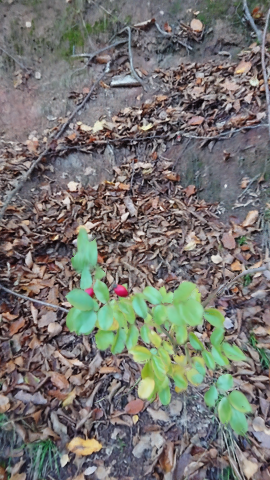}
  \\
  \tiny
  Canonical & \tiny Dampening $0.3\times$ & \tiny Normal $1\times$ & \tiny Exaggeration $2\times$ & \tiny Exaggeration $3\times$ & \tiny Exaggeration $5\times$
  \\
  
  \includegraphics[angle=90,origin=c,width=.165\columnwidth]{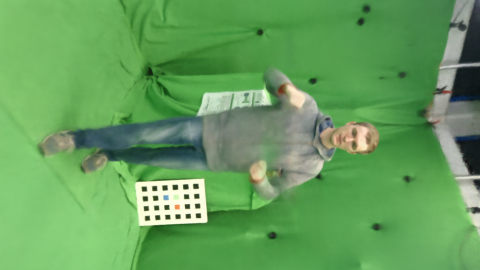}
  &
  \includegraphics[angle=90,origin=c,width=.165\columnwidth]{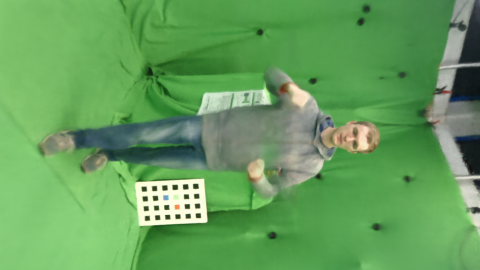}
  &
  \includegraphics[angle=90,origin=c,width=.165\columnwidth]{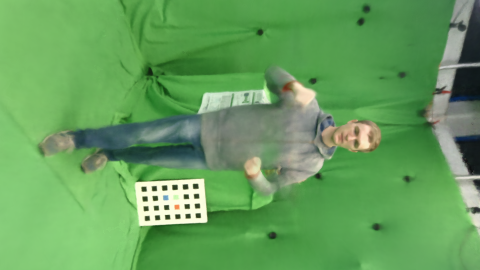}
  &
  \includegraphics[angle=90,origin=c,width=.165\columnwidth]{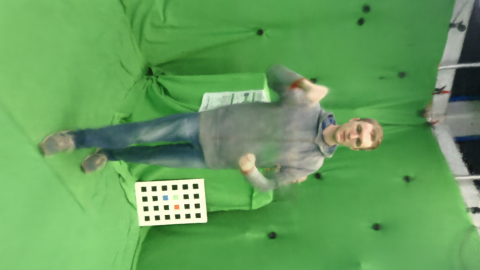}
  &
  \includegraphics[angle=90,origin=c,width=.165\columnwidth]{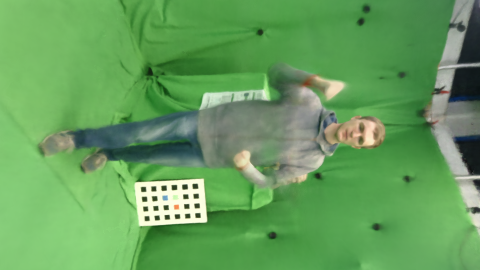}
    &
  \includegraphics[angle=90,origin=c,width=.165\columnwidth]{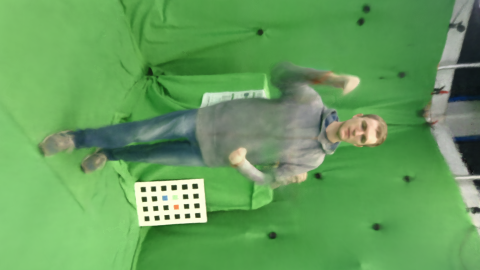}
  \\
  \tiny
  Canonical & \tiny Dampening $0.2\times$ & \tiny Dampening $0.5\times$ & \tiny Normal $1\times$ & \tiny Exaggeration $1.15\times$ & \tiny Exaggeration $2\times$

\end{tabular}
  \caption{We exaggerate or dampen the motion relative to the canonical model, and render the result into a novel view.}
  \label{fig:exaggeration}
\end{figure}

\subsection{Deformation Exaggeration/Dampening} 
We can manipulate the deformation even further.
Specifically, We can exaggerate or dampen deformations relative to the canonical model by scaling all offsets with a constant $m\in\mathbb{R}$: $(\mathbf{c}, o)=\mathbf{v}(\mathbf{x} + m\mathbf{b}(\mathbf{x},\mathbf{l}_i))$.
Fig.~\ref{fig:exaggeration} contains examples.

\begin{figure}
  \includegraphics[width=.115\textwidth]{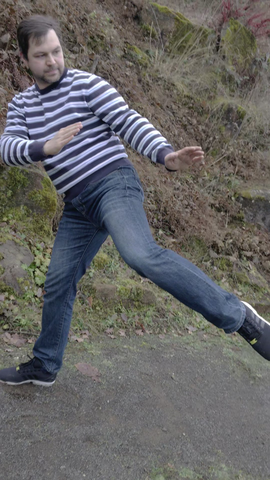}%
  \includegraphics[width=.115\textwidth]{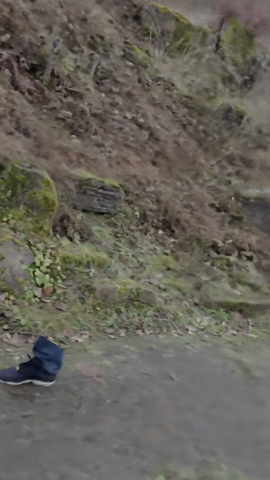} 
  \includegraphics[width=.115\textwidth]{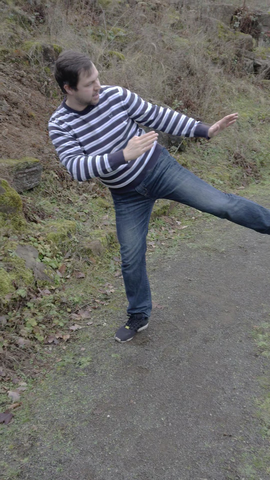}%
  \includegraphics[width=.115\textwidth]{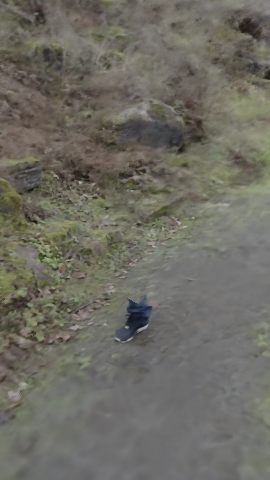}
  
  \includegraphics[width=.23\textwidth]{latex/figures/foreground_removal/stick_still_2_groundtruth_image00174.png}%
  \includegraphics[width=.23\textwidth]{latex/figures/foreground_removal/stick_still_2_f4_ray5x64_moivd600_mdiod30_L1rig2_003_rgb_000173.png} 
  \caption{(Left) the groundtruth input image and (right) a rendering without non-rigid foreground. %
  }
  \label{fig:foreground_removal_additional}
\end{figure}

\subsection{Forced Background Stabilization}
Since we do not require any pre-computed foreground-background segmentation, NR-NeRF has to assign rigidity scores without supervision.
Occasionally, this insufficiently constrains the background and leads to small motion.
We can fix this in some cases by enforcing a stable background at test time: 
we set the regressed score to 0 if it is below some threshold $r_{\textit{min}}$.
If the rigid background has sufficiently small scores assigned to it relative to the non-rigid part of the scene, this forces the background to remain static for all time steps and views.
For results, we refer to the supplemental video.

\section{Limitations}\label{sec:limitations}

\begin{figure}

  \includegraphics[width=.15\textwidth]{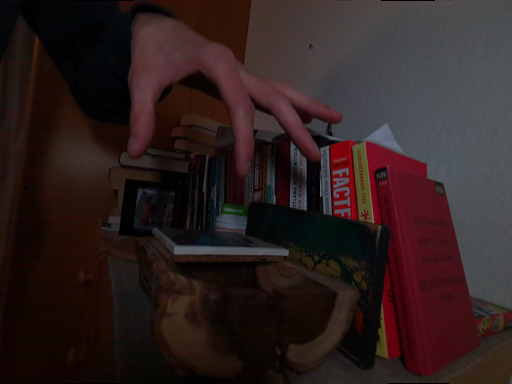}
  \includegraphics[width=.15\textwidth]{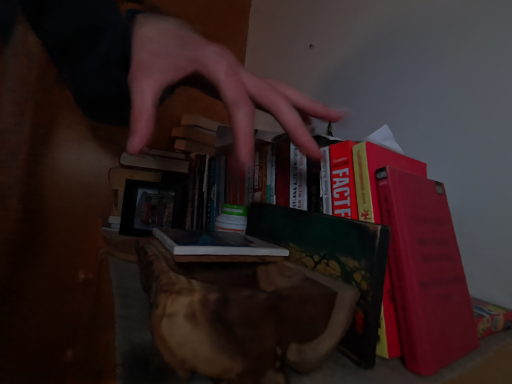}
  \includegraphics[width=.15\textwidth]{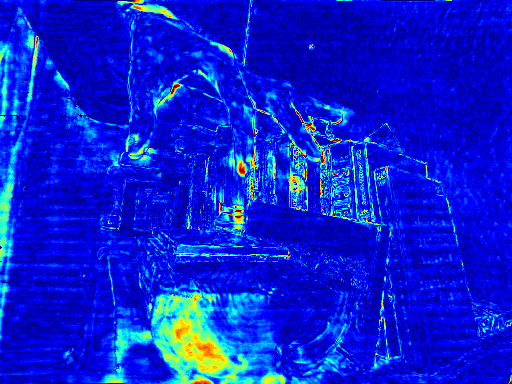}
  \caption{The input (left) is reconstructed by NR-NeRF (middle). The bottom of the image exhibits local shadowing absent at other time steps, which leads to a high reconstruction error (right).}
  \label{fig:error_maps}
  
\end{figure}

We do not account for appearance changes that are due to deformation or lighting changes.
For example, temporally changing shadowing in the input images is an issue, as Fig.~\ref{fig:error_maps} demonstrates.

Foreground removal can fail if a part of the foreground is entirely static (\textit{e.g.,} the foot in Fig.~\ref{fig:foreground_removal_additional}).

\section{Additional Comparisons}\label{sec:additional_comparisons}

\begin{figure*}
\setlength{\tabcolsep}{0.1em} %
\def\arraystretch{0.5} %
\begin{tabular}{cccccc}

  Input & Ours & NSFF \cite{li2020NSFF} & Na\"ive NR-NeRF & Rigid (no view dep.) & Rigid (view dep.)\\

  \includegraphics[width=.165\textwidth]{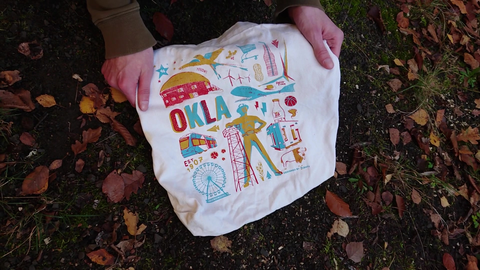}
  &
  \includegraphics[width=.165\textwidth]{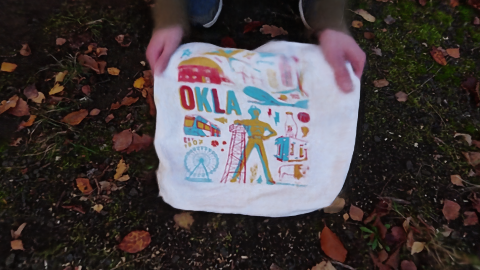}
  &
  \includegraphics[width=.165\textwidth]{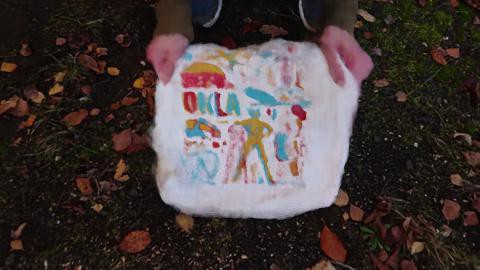}
  &
  \includegraphics[width=.165\textwidth]{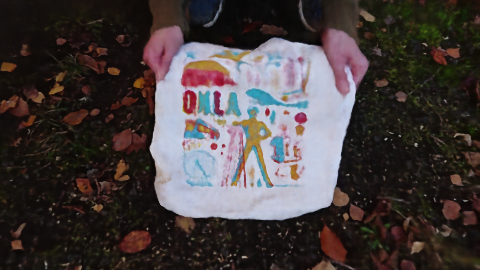}
  &
  \includegraphics[width=.165\textwidth]{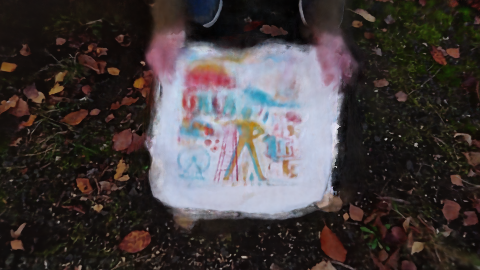}
    &
  \includegraphics[width=.165\textwidth]{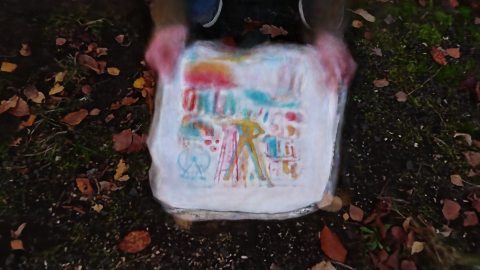}
  \\
  
  \includegraphics[width=.165\textwidth]{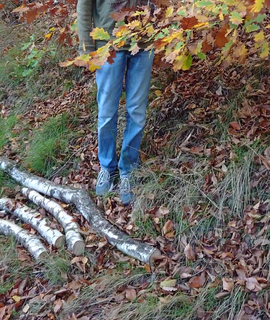}
  &
  \includegraphics[width=.165\textwidth]{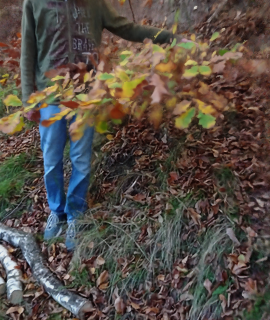}
  &
  \includegraphics[width=.165\textwidth]{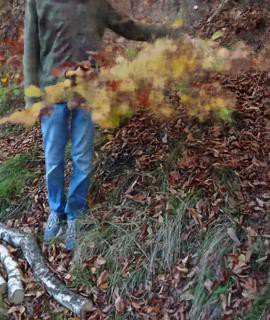}
  &
  \includegraphics[width=.165\textwidth]{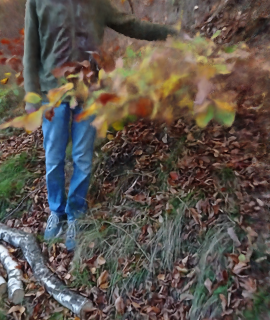}
  &
  \includegraphics[width=.165\textwidth]{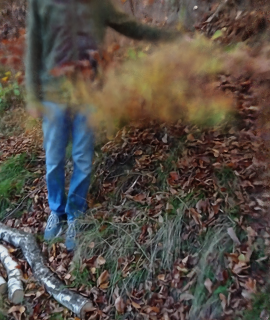}
    &
  \includegraphics[width=.165\textwidth]{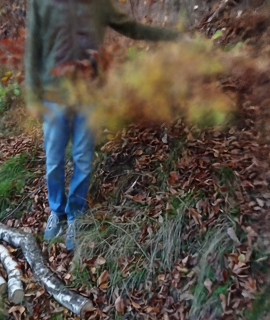}
  \\

  \includegraphics[width=.165\textwidth]{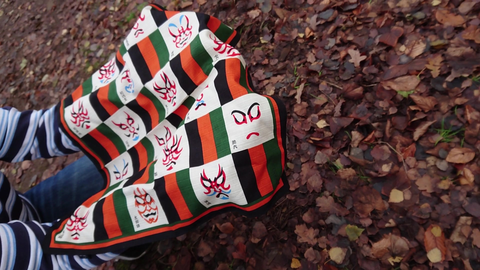}
  &
  \includegraphics[width=.165\textwidth]{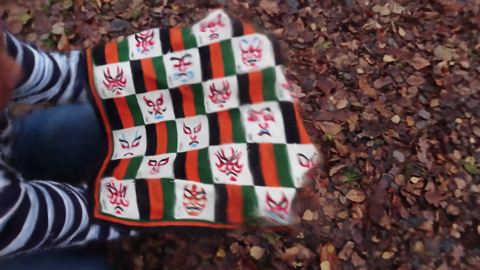}
  &
  \includegraphics[width=.165\textwidth]{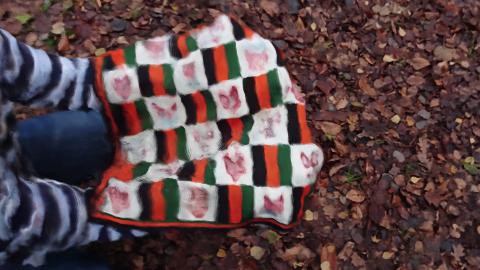}
  &
  \includegraphics[width=.165\textwidth]{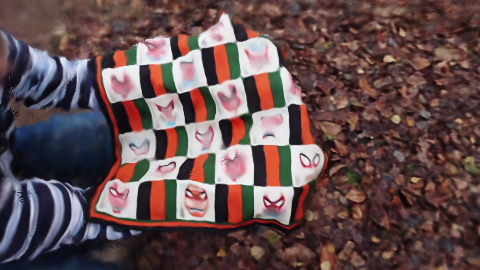}
  &
  \includegraphics[width=.165\textwidth]{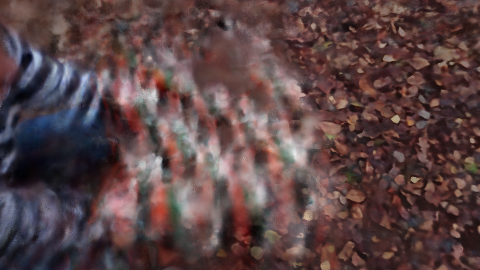}
    &
  \includegraphics[width=.165\textwidth]{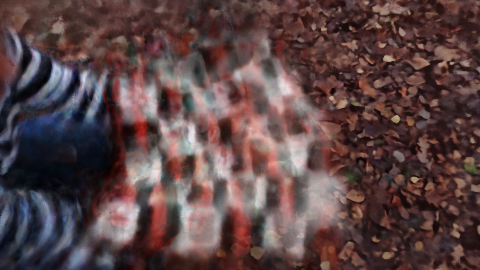}
  \\
  
  \includegraphics[width=.165\textwidth]{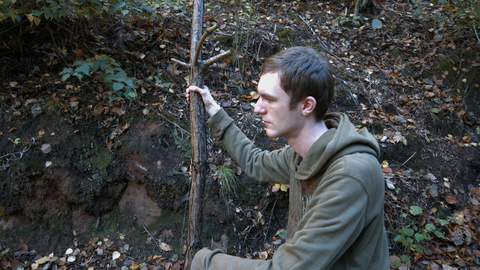}
  &
  \includegraphics[width=.165\textwidth]{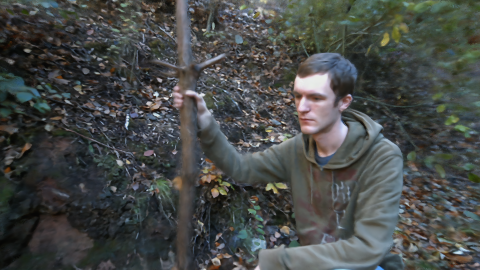}
  &
  \includegraphics[width=.165\textwidth]{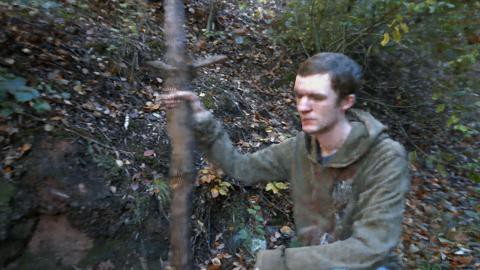}
  &
  \includegraphics[width=.165\textwidth]{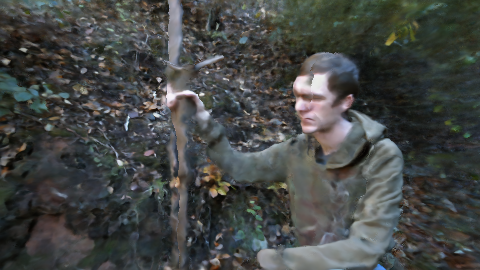}
  &
  \includegraphics[width=.165\textwidth]{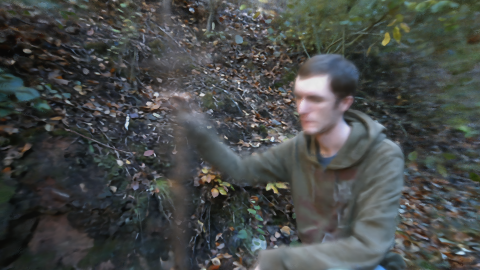}
    &
  \includegraphics[width=.165\textwidth]{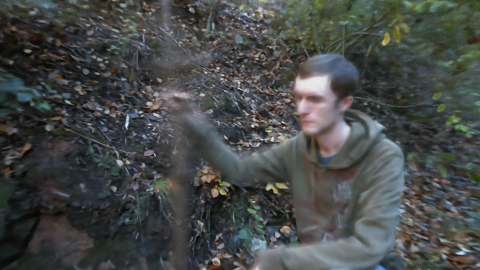}
  \\

\end{tabular}
  \caption{Under challenging novel view scenarios, our method benefits from the geometry and appearance information that the canonical volume has accumulated from all time steps, which allows NR-NeRF to output sharp results. 
  Both the concurrent, non-peer-reviewed Neural Scene Flow Fields~\cite{li2020NSFF} and na\"ive NR-NeRF however entangle deformation with geometry and appearance by conditioning the 'canonical' volume on a time-dependent deformation latent code. 
  This makes sharing information across time more difficult, leading to blurrier results in challenging novel view scenarios compared to NR-NeRF's results. 
  Finally, rigid NeRF shows a blurry mix of the deformations observed over the entire input sequence, which highlights the need to account for deformations in the scene.}
  \label{fig:additional_comparisons_nsff}
\end{figure*}

We show some preliminary qualitative comparisons to the concurrent, non-peer-reviewed work Neural Scene Flow Fields~\cite{li2020NSFF} in Fig.~\ref{fig:additional_comparisons_nsff}.

{\small
\bibliographystyle{ieee_fullname}
\bibliography{main_doc}
}

\end{document}